\documentclass[review]{elsarticle}

\usepackage{hyperref}

\journal{Journal of \LaTeX\ Templates}

\biboptions{sort&compress,comma}
\usepackage{epstopdf}
\usepackage{amsfonts}
\usepackage{color}
\usepackage{enumitem}
\usepackage{url}
\usepackage{graphicx}
\usepackage{diagbox}
\usepackage{array}
\usepackage{amsmath,amsthm,amssymb}
\usepackage{verbatim}
\usepackage{multirow}

\usepackage{algorithm}  
\usepackage{algorithmicx} 
\usepackage{algpseudocode}  
\usepackage{caption}

\makeatletter

\newcounter{phase}[algorithm]
\newlength{\phaserulewidth}
\newcommand{\phaseTitle}{Stage}

\newcommand{\phase}[1]{%
  \vspace{-5ex}
  \Statex\leavevmode\llap{\rule{\dimexpr\labelwidth+\labelsep}{\phaserulewidth}}\rule{\linewidth}{\phaserulewidth}
  \Statex\strut\refstepcounter{phase}\item[\textbf{\phaseTitle~\thephase:~}\textit{~#1}] 
  \vspace{-3ex}\Statex\leavevmode\llap{\rule{\dimexpr\labelwidth+\labelsep}{\phaserulewidth}}\rule{\linewidth}{\phaserulewidth}}

\newtheorem{proposition}{Proposition}

\begin{document}

\begin{frontmatter}

\title{Autoencoder Based Sample Selection \\for Self-Taught Learning}

\author[mymainaddress]{Siwei Feng}
\ead{siwei@umass.edu}

\author[yuaddr1,yuaddr2,yuaddr3]{Han Yu}
\ead{han.yu@ntu.edu.sg}

\author[mymainaddress]{Marco F. Duarte\corref{mycorrespondingauthor}}
\cortext[mycorrespondingauthor]{Corresponding author}
\ead{mduarte@ecs.umass.edu}

\address[mymainaddress]{Department of Electrical and Computer Engineering, University of Massachusetts, Amherst, MA 01003 USA}
\address[yuaddr1]{School of Computer Science and Engineering, Nanyang Technological University, Singapore 639798}
\address[yuaddr2]{Joint NTU-UBC Research Centre of Excellence in Active Living for the Elderly, Singapore 639798}
\address[yuaddr3]{Alibaba-NTU Singapore Joint Research Institute, Singapore 639798}

\begin{abstract}
Self-taught learning is a technique that uses a large number of unlabeled data as source samples to improve the task performance on target samples. Compared with other transfer learning techniques, self-taught learning can be applied to a broader set of scenarios due to the loose restrictions on the source data. However, knowledge transferred from source samples that are not sufficiently related to the target domain may negatively influence the target learner, which is referred to as negative transfer. In this paper, we propose a metric for the relevance between a source sample and the target samples. To be more specific, both source and target samples are reconstructed through a single-layer autoencoder with a linear relationship between source samples and reconstructed target samples being simultaneously enforced. An $\ell_{2,1}$-norm sparsity constraint is imposed on the transformation matrix to identify source samples relevant to the target domain. Source domain samples that are deemed relevant are assigned pseudo-labels reflecting their relevance to target domain samples, and are combined with target samples in order to provide an expanded training set for classifier training. Local data structures are also preserved during source sample selection through spectral graph analysis. Promising results in extensive experiments show the advantages of the proposed approach.
\end{abstract}


\end{frontmatter}


\section{Introduction}

Supervised learning is widely used in many machine learning tasks \cite{dietterich1998approximate,cohen2013applied, hong2018multi}. However, applications of supervised learning methods are limited in practical scenarios due to its requirements on large-scale labeled training datasets\footnote{``Training data" is used in the sequel to denote data used for model learning.} with both training and testing data sharing the same label and feature space, which lead to high costs in collecting eligible training data \cite{russakovsky2015imagenet, rozantsev2018beyond}.\par
Several techniques have been proposed to tackle the limitations of supervised learning methods. Semi-supervised learning algorithms \cite{zhu2006class, nigam2000text} use both labeled and unlabeled data to improve performance when labeled training data are limited. However, the success of many semi-supervised learning algorithms highly depends on the validity of assumptions that the unlabeled and labeled data have the same distribution \cite{zhu2006class} or class labels \cite{nigam2000text}. Therefore, it is still difficult to gather unlabeled data that satisfy these preconditions. \par
In order to further loosen the restrictions on training data, many transfer learning approaches \cite{pan2010survey} have been proposed to use the knowledge obtained from auxiliary domains\footnote{``Source domain'' is used in the sequel to denote auxiliary domains that are used to improve target domain task performance.} to improve the performance on target domain tasks. \emph{Self-taught learning} \cite{raina2007self, dai2008self, kuen2015self, bazzani2016self, jie2017deep, kemker2017self, he2017self, wang2013robust, li2018self} is a type of transfer learning technique that employs unlabeled auxiliary data to improve the performance of a supervised learning task when labeled training data are limited. Though similar to semi-supervised learning, self-taught learning methods have fewer restrictions on unlabeled data, as they allow the label spaces and marginal probability distributions of unlabeled and labeled data to be different. In self-taught learning, unlabeled data are used as the source from which the knowledge learned is applied to tasks performed on labeled target data. Such a loose restriction on unlabeled data significantly simplifies learning due to the huge volume of unlabeled data we can access. However, the easily obtained unlabeled data inevitably contain samples that are only weakly related to the labeled training data, which may even harm the supervised learning performance if we treat them equally as other unlabeled samples during knowledge transfer. Target task performance degradation due to knowledge transfer from source domains that are not sufficiently related is known as \emph{negative transfer} \cite{rosenstein2005transfer}, which has been studied under several transfer learning scenarios \cite{duan2012exploiting, yang2015learning}. However, to the best of our knowledge, the problem of negative transfer in self-taught learning has not been studied before. \par

\begin{figure*}[t]
\centering
\includegraphics[width=13cm]{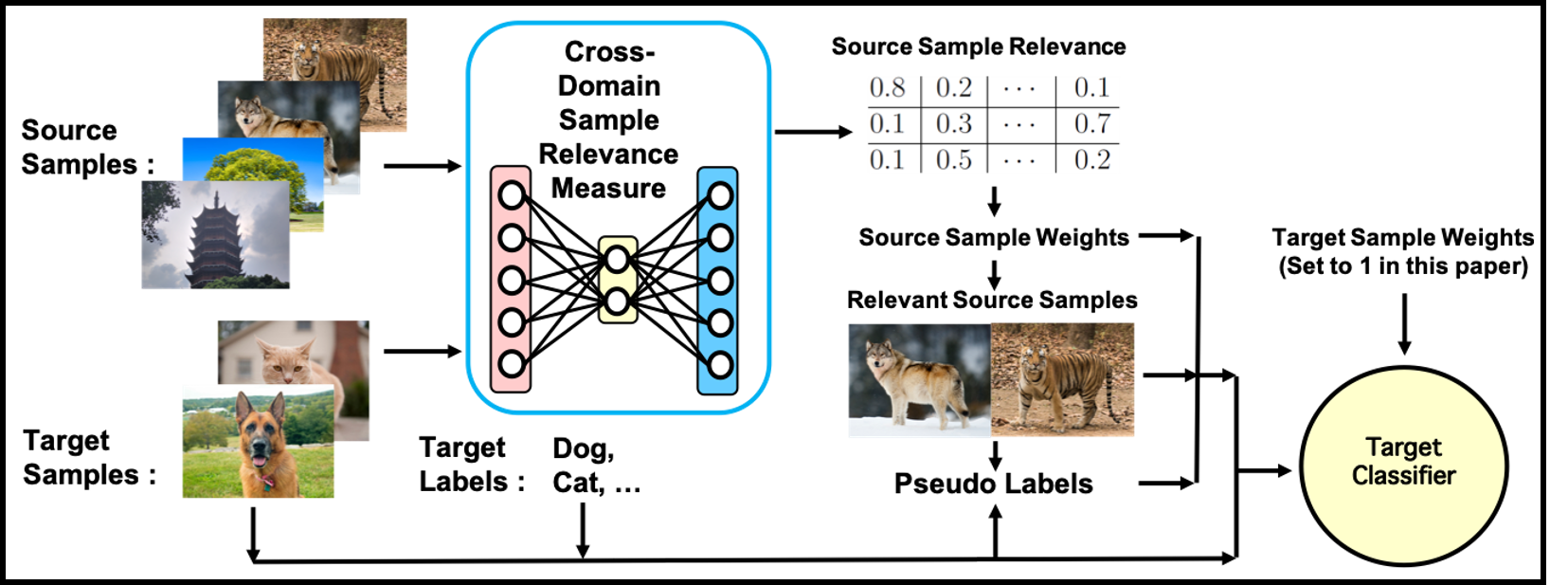}
\caption{Block diagram for the GASTL framework.}
\label{framework}
\end{figure*}

In this paper, we propose a novel algorithm for self-taught learning with unlabeled source data that are related to labeled target data to be selected with the purpose of reducing negative transfer. The algorithm leverages a linear mapping, a $k$-nearest neighbor ($k$NN) graph, and a single-layer autoencoder to obtain a metric for cross-domain relevance. We refer to this method as graph and autoencoder-based self-taught learning (GASTL). The GASTL framework includes two modules: a source sample re-weighting module and a classifier training module. In the first module, we assign each unlabeled source sample a weight that indicates its relevance to labeled target samples.\footnote{The setting of self-taught learning requires source samples to be unlabeled and target samples to be labeled. Therefore we do not specify the availability of label information for both source and target samples in the sequel.} In the second module, source samples with large weights are selected as a transfer training set to be combined with target data to train a classifier. Each selected source sample is assigned a pseudo-label from the target domain label space to be used during classifier training. The source sample weights are also used during classifier training. The trained classifier is then used to predict labels of unseen target samples. Figure \ref{framework} shows the flowchart of GASTL.  \par
The key contributions of this paper are as follows:
\begin{itemize}
\item We propose a novel metric for the relevance of each source sample to the target domain in the scenario of self-taught learning based on an autoencoder and graph data regularization. To the best of our knowledge, we are the first to measure cross-domain sample relevance in order to tackle the issue of negative transfer in self-taught learning problems.
\item We propose a novel classifier training scheme with both selected source samples and target samples as the training dataset with the relevance of each source sample to the target domain being considered. We are not aware of existing self-taught learning approaches that integrate cross-domain relevance into classifier training.
\item We present an efficient solver for the knowledge transfer optimization problem described above that relies on an iterative scheme based on gradient descent of the proposed objective function. 
\item We provide multiple numerical results to demonstrate the performance improvements in terms of classification accuracy and sensitivity to parameters achieved by the proposed method compared with state-of-the-art self-taught learning methods and other relevant techniques.
\end{itemize}
The rest of this paper is organized as follows. Section \ref{notations} introduces notation. Section \ref{rela} overviews relevant techniques. The proposed framework is presented in Section \ref{method}. Experimental results are provided in Section \ref{exp}. Section \ref{conclusion} concludes with suggestions for future work.

\section{Notation}
\label{notations}

Vectors are denoted by bold lowercase letters while matrices are denoted by bold uppercase letters. The superscript $T$ of a matrix denotes the transposition operation. For a matrix $\mathbf{A}$, $\mathbf{A}^{(q)}$ denotes the $q^{\mathrm{th}}$ column and $\mathbf{A}_{(p)}$ denotes the $p^{\mathrm{th}}$ row, while $\mathbf{A}^{(p,q)}$ denotes the entry at the $p^{\mathrm{th}}$ row and $q^{\mathrm{th}}$ column. The $\ell_{r,p}$-norm for a matrix $\mathbf{W} \in \mathbb{R}^{a \times b}$ is denoted as $\|\mathbf{W}\|_{r,p} = \left( \sum_{i=1}^a \left( \sum_{j=1}^b |\mathbf{W}^{(i,j)}|^r \right)^{p/r} \right)^{1/p}$. The $\ell_r$-norm for a vector $\mathbf{w} \in \mathbb{R}^a$ is denoted as $\| \mathbf{w} \|_r = \left( \sum_{i=1}^a | \mathbf{w}^{(i)} |^r \right)^{1/r}$.
The trace of a matrix $\textbf{L} \in \mathbb{R}^{a \times a}$ is defined as ${\rm{Tr}} (\mathbf{L}) = \sum_{i=1}^a \mathbf{L}^{(i,i)}$. We use $\mathbf{1}$ and $\mathbf{0}$ to denote an all-ones and all-zeros matrix or vector of the appropriate size, respectively. 
We use $\mathbf{X} = [\mathbf{X}^{(1)}, \mathbf{X}^{(2)}, \cdots , \mathbf{X}^{(n)}] \in \mathbb{R}^{d \times n}$ to denote a sample set, where $\mathbf{X}^{(i)} \in \mathbb{R}^d$ is the $i^{\mathrm{th}}$ sample in $\mathbf{X}$ for $i = 1, 2, \cdots , n$, and where $d$ and $n$ denote the sample dimensionality and the number of samples in $\mathbf{X}$, respectively. \par
In transfer learning, we use $\mathcal{D}$ to denote a domain and $\mathcal{T}$ to denote a task. A domain $\mathcal{D}$ consists of a feature space $\mathcal{X}$ and a marginal probability distribution $P(\mathbf{X})$ over a sample set $\mathbf{X}$. A task $\mathcal{T}$ consists of a label space $\mathcal{Y}$ and an objective predictive function $f(\mathbf{X}, \mathbf{Y})$ to predict the corresponding labels $\mathbf{Y}$ of a sample set $\mathbf{X}$. We refer readers to \cite{pan2010survey} for a detailed explanation of these notations. We use $\mathcal{D}_{\rm src} = \{ \mathcal{X}_{\rm src}, P(\mathbf{X}_{\rm src}) \}$ and $\mathcal{T}_{\rm src} = \{ \mathcal{Y}_{\rm src}, f(\mathbf{X}_{\rm src}, \mathbf{Y}_{\rm src}) \}$ to denote the source domain and task, and use $\mathcal{D}_{\rm trg} = \{ \mathcal{X}_{\rm trg}, P(\mathbf{X}_{\rm trg}) \}$ and $\mathcal{T}_{\rm trg}= \{ \mathcal{Y}_{\rm trg}, f(\mathbf{X}_{\rm trg}, \mathbf{Y}_{\rm trg}) \}$ for the target domain and task. \par

\section{Background and Related Work}
\label{rela}

\subsection{Self-Taught Learning}
\label{stlintro}

Transfer learning methods can be classified as homogeneous ($\mathcal{X}_{\rm src} = \mathcal{X}_{\rm trg}$) or heterogeneous ($\mathcal{X}_{\rm src} \neq \mathcal{X}_{\rm trg}$), and as transductive ($\mathcal{T}_{\rm src} = \mathcal{T}_{\rm trg}$) or inductive ($\mathcal{T}_{\rm src} \neq \mathcal{T}_{\rm trg}$) \cite{pan2010survey}. The self-taught learning setting assumes different label spaces between the source and target domains, and label information is assumed to be unavailable in the source domain. Therefore, the self-taught learning setting is similar to the inductive transfer learning setting when no labeled data is available in the source domain. The goal of self-taught learning is to use the unlabeled source domain data to help improve the target domain task performance. In this paper we focus on homogeneous self-taught learning. \par
The idea of self-taught learning was first proposed by Raina et al.\ \cite{raina2007self}.\footnote{We use abbreviation STL to denote the method of \cite{raina2007self} in the sequel, while we use the full name ``self-taught learning`` for the class of learning problems.} In STL, a dictionary $\mathbf{D}$ is first learned using source domain samples. After that, the sparse coefficients for target domain sampels $\mathbf{A}_{\rm trg}$ are computed based on $\mathbf{D}$. Finally, a classifier is learned on $\mathbf{A}_{\rm trg}$ as well as target domain labels by applying a supervised learning algorithm. Robust and discriminative self-taught learning (RDSTL) \cite{wang2013robust} is an extension of STL that takes advantage of the label information of target samples during learning and makes the dictionary learning process more robust to noise and outliers by imposing an $\ell_{2,1}$-norm constraint on both the dictionary learning reconstruction loss and the sparse coefficient matrix. Self-taught low-rank (S-Low) coding \cite{li2018self} is suitable for both clustering and classification tasks in visual learning. By imposing a low-rank constraint onto the sparse coefficient matrix, S-Low coding is able to characterize the global structure information in the target domain. Self-taught clustering \cite{dai2008self} is the first algorithm proposed to tackle unsupervised inductive transfer learning problems and aims at clustering a small amount of target unlabeled data by learning a useful feature representation with the help of large amounts of unlabeled source domain data. Kuen et al.\ \cite{kuen2015self} incorporates self-taught learning into visual tracking, where local representations are learned offline on unlabeled data and transferred to the observational model of the proposed tracker. Kemker et al.\ \cite{kemker2017self} applies self-taught learning to hyperspectral image classification by learning models from sufficiently large quantities of unlabeled source data that are distinct from the labeled target data to extract generalizable features. The trained models are then used to extract features on the labeled target data for classification. He et al.\ \cite{he2017self} combines self-taught learning, sparse autoencoder, and radial basis functions to the field of wound infection detection to solve the problem of insufficient labeled wound infection samples. A basis vector is first learned on the easily obtained unlabeled pollutant gas samples. Then new representation of wound infection samples are learned using the basis vector under a sparsity constraint for classification. The idea of self-taught learning has also been incorporated into object localization \cite{bazzani2016self, jie2017deep} to learn object detectors without or with little human supervision. \par
Although these self-taught learning approaches use different schemes for knowledge transfer, they all use the entire source sample set without considering their relevance to the target domain, which makes these methods potentially vulnerable to negative transfer. \par

\subsection{Single-Layer Autoencoder}
\label{ae}

A single-layer autoencoder is an artificial neural network that aims to reconstruct inputs by using only a single hidden layer, and is widely used in transfer learning \cite{zhu2018transfer}. Given input data $\mathbf{x} \in \mathbb{R}^d$, an autoencoder first maps $\mathbf{x}$ to a compressed data representation $\mathbf{z} \in \mathbb{R}^m$ in a hidden layer, given by $\mathbf{z} = {\rm f} ( \mathbf{W_1x} + \mathbf{b_1} )$, where $\mathbf{W_1} \in \mathbb{R}^{m \times d}$ is a weight matrix, $\mathbf{b_1} \in \mathbb{R}^m$ is a bias vector, and ${\rm f} (\cdot)$ is an elementary nonlinear activation function. This part is referred to as an \textit{encoder}. The second step is to map the compressed data representation $\mathbf{z}$ to output data $\mathbf{\bar{\mathbf{x}} \in \mathbb{R}^d}$, which is $\mathbf{\bar{\mathbf{x}}} = {\rm g} ( \mathbf{W_2z} + \mathbf{b_2} )$, where $\mathbf{W_2} \in \mathbb{R}^{d \times m}$ and $\mathbf{b_2} \in \mathbb{R}^d$ are the corresponding weight matrix and bias vector, respectively. This part is referred to as a \textit{decoder}. \par
The optimization problem underlying autoencoder training is to minimize the difference between the input data and the output data. To be more specific, given a set of data $\mathbf{X} = [\mathbf{X}^{(1)}, \mathbf{X}^{(2)}, \cdots, \mathbf{X}^{(n)} ]$, the parameters $\mathbf{W_1}$, $\mathbf{W_2}$, $\mathbf{b_1}$, and $\mathbf{b_2}$ are adapted to minimize the reconstruction error $\sum_i \| \mathbf{X}^{(i)} - \mathbf{\bar{X}}^{(i)} \|_2^2$, where $\mathbf{\bar{X}}^{(i)}$ is the output of autoencoder to the input $\mathbf{X}^{(i)}$.

\subsection{Instance-Based Transfer Learning}

Instance-based transfer learning assumes that certain instances of data from the source domain can be reused for learning in the target domain by reweighting. Instance reweighting and sampling based on instance importance are two major techniques in this context \cite{pan2010survey}. Kernel mean matching (KMM) \cite{huang2007correcting} and multiscale landmarks selection (MLS) \cite{aljundi2015landmarks} are two instance-based transfer learning methods that fit our setting of unlabeled source data since neither of them require label information from the source domain. KMM is a nonparametric method which directly produces resampling weights without requiring the estimation of biased densities or selection probabilities, or the assumption that the probabilities of the different classes are known. MLS selects landmarks that are similarly distributed in the two domains to reduce the discrepancy between the domains, where each instance from the union of source and target domain data is considered as a candidate landmark, and the candidate is kept as a landmark if its quality is sufficiently high according to some measure.

\section{Proposed Method}
\label{method}

In this section, we introduce our proposed GASTL approach. The basic framework of GASTL is to reconstruct both source and target samples through a single-layer autoencoder, while simultaneously enforcing a linear relationship between source samples and target samples. Both global and local data structures are preserved through a single-layer autoencoder and spectral graph analysis, respectively. We develop a metric for the relevance between each source sample and the target samples, which is used for source sample selection for knowledge transfer. Meanwhile, a weight is assigned to each source sample reflecting its relevance to the target samples for the subsequent classifier training, during which each selected source sample is assigned a pseudo-label from the target domain label space and combined with target samples to build the classifier training sample set. Source sample weights are also considered during classifier training. Finally, the trained classifier is used to predict labels of unseen target samples.

\subsection{Knowledge Transfer and Relevance Measure}

In this section, we present the problem formulation of our knowledge transfer scheme. We also propose a measure for relevance between each source sample and the target samples.

\subsubsection{Objective Function}
\label{obj}

The objective function of GASTL includes four parts: a data reconstruction term, a domain mapping term, a regularization term for sample selection, and a term based on spectral graph analysis for local data structure preservation. The details of these four terms are described below. \par
Many transfer learning methods perform knowledge transfer from a source domain to a target domain by finding a mapping between them \cite{sanodiya2019framework, hu2016multi}. In particular, we can set ${\rm h_1} (\mathbf{X}_{\rm trg}) = {\rm h_2} (\mathbf{X}_{\rm src})\mathbf{A}$, where ${\rm h_1} (\cdot)$ and ${\rm h_2} (\cdot)$ are two transformations, while $\mathbf{A}$ is a matrix that linearly maps transformed source samples ${\rm h_2} (\mathbf{X}_{\rm src})$ into transformed target samples ${\rm h_1} (\mathbf{X}_{\rm trg})$. More specifically, the mapping is obtained from the optimization:
\begin{equation*}
\label{generalReconst}
\small
\min_{\mathbf{\Theta}, \mathbf{A}} \mathcal{M} (\mathbf{\Theta}, \mathbf{A}) + \lambda \mathcal{R} (\mathbf{A}),
\end{equation*}
where  $\mathbf{\Theta}$ is a set of parameters used for the nonlinear mappings ${\rm h_1}$ and ${\rm h_2}$, while $\mathcal{M} (\mathbf{\Theta}, \mathbf{A}) = \mathcal{L} \left( {\rm h_1} (\mathbf{X}_{\rm trg}), {\rm h_2} (\mathbf{X}_{\rm src}) \mathbf{A} \right)$  denotes a cost function for domain mapping, where $\mathcal{L} (\cdot, \cdot)$ is a loss function and $\mathcal{R} (\cdot)$ corresponds to a regularization function on $\mathbf{A}$ to avoid overfitting.\footnote{We empirically found that regularizing $\mathbf{\Theta}$ did not noticeably affect the performance of knowledge transfer. Therefore, we do not pursue such regularization in this paper.} \par
A simple way to achieve domain mapping is to assume a linear mapping between source and target data, which is $\mathbf{X}_{\rm trg} = \mathbf{X}_{\rm src}\mathbf{A}$. This requires the cost $\mathcal{M} (\mathbf{\Theta}, \mathbf{A}) = \mathcal{L} (\mathbf{X}_{\rm trg}, \mathbf{X}_{\rm src} \mathbf{A})$. The use of a linear mapping in knowledge transfer is often computationally efficient. However, the success of this knowledge transfer scheme relies on an assumption that $\mathbf{X}_{\rm trg} \in {\rm span} (\mathbf{X}_{\rm src})$ \cite{shao2014generalized}. Due to the ubiquitous large discrepancy between the source and target domains in self-taught learning scenarios, $\mathbf{X}_{\rm trg}$ is usually not in the span of $\mathbf{X}_{\rm src}$, and hence a linear reconstruction scheme can hardly do well in knowledge transfer. Therefore, we aim to find a non-linear reconstruction scheme that can decrease the discrepancy between source and target domains. One possible way to do this is to find a nonlinear transformation on $\mathbf{X}_{\rm trg}$, and recover the output of this transformation as a linear transformation of source samples which are relevant to the target samples. That is, ${\rm h} (\mathbf{X}_{\rm trg}) = \mathbf{X}_{\rm src}\mathbf{A}$, where ${\rm h} (\cdot)$ is a nonlinear transformation. Furthermore, due to the possible large diversity of source samples compared with the target samples, we can assume that the feature space shared by both source and target domains are separated into several clusters: the source samples lie near a union of many clusters, while the target samples concentrate near a single cluster. Intuitively, negative transfer can be alleviated by using source samples that are close to target samples for knowledge transfer. \par
As mentioned in Section \ref{ae}, a single-layer autoencoder aims at minimizing the reconstruction error between output and input data. We use $\mathbf{X} = [\mathbf{X}_{\rm src} \ \mathbf{X}_{\rm trg}]$ as the input to a single-layer autoencoder by optimizing a reconstruction error-driven loss function:
\begin{equation}
\label{reconErr}
\small
\mathcal{L}(\mathbf{\Theta}) = \frac{1}{2n} \| \mathbf{X} - {\rm h} (\mathbf{X};\mathbf{\Theta}) \|_F^2,
\end{equation}
where $n = n_{\rm src} + n_{\rm trg}$, $\mathbf{\Theta} = [\mathbf{W_1}, \mathbf{W_2}, \mathbf{b_1}, \mathbf{b_2}]$, and ${\rm h} (\mathbf{X};\mathbf{\Theta}) = \  {\rm g} ( \mathbf{W_2} \cdot {\rm f} ( \mathbf{W_1}\mathbf{X} + \mathbf{b_1} )  + \mathbf{b_2} )$.\footnote{We often drop the dependence on $\mathbf{\Theta}$ for readability, i.e.\ we use ${\rm h} (\mathbf{X})$ to denote ${\rm h} (\mathbf{X}; \mathbf{\Theta})$ when no ambiguity is caused.} 
We use the sigmoid function as the activation function: ${\rm f}(z) = {\rm g} (z) = 1/(1+ {\rm{exp}} (-z))$. In  \eqref{reconErr}, both source and target samples share the same parameters to train an autoencoder, which makes the reconstructed source and target samples lie in the same submanifold under the learned parameters $\mathbf{\Theta}$. Meanwhile, we use the following minimization problem for the purpose of domain mapping:
\begin{equation}
\label{crossdomainReconst}
\small
\mathcal{C}(\mathbf{\Theta}, \mathbf{A}) = \frac{1}{2n_{\rm trg}} \| \mathbf{X}_{\rm src}\mathbf{A} - {\rm h} (\mathbf{X}_{\rm trg}; \mathbf{\Theta}) \|_F^2.
\end{equation}
That is, we enforce the target samples in the autoencoder output to be reconstructed by a linear combination of the source samples. While it is feasible to separate the optimization of  \eqref{reconErr} and  \eqref{crossdomainReconst}, we observed that a joint framework is able to provide better knowledge transfer performance. Due to the nonlinear nature of transformation featured by a single-layer autoencoder, the distribution gap can be ameliorated through minimizing $\mathcal{C}(\mathbf{\Theta}, \mathbf{A})$ with respect to $\mathbf{\Theta}$ and $\mathbf{A}$. Therefore, we define the mapping cost 
\begin{equation*}
\small
\mathcal{M} (\mathbf{\Theta}, \mathbf{A}) = \mathcal{L} (\mathbf{\Theta}) + \mu \mathcal{C} (\mathbf{\Theta}, \mathbf{A}),
\label{mapping}
\end{equation*}
where $\mu$ is a balance parameter, to obtain a nonlinear mapping between source samples and target samples.\par 
Since each row of $\mathbf{A}$ indicates the importance of the corresponding source sample in reconstructing transformed target samples, we use the $\ell_2$-norm of each row of $\mathbf{A}$ to measure the \textit{relevance} between a source sample and the target samples. This leads to an $\ell_{2,1}$-norm regularization function $\mathcal{R}(\mathbf{A}) = \|\mathbf{A}\|_{2,1}$ that enforces row sparsity on the transformation matrix $\mathbf{A}$. 

Data transformations based on autoencoders only guarantee broad data structure preservation, which does not take pair-wise relationships between data points into consideration. Therefore, we need to include local geometric structures from the data into our objective function. Local geometric structures of the data often contain discriminative information of neighboring data point pairs~\cite{cai2010unsupervised, shang2016subspace, shang2017non, shang2019unsupervised, yu2014learning, hong2014image}, in which nearby data points are assumed to have similar representations. In order to characterize the local data structure, we construct a $k$-nearest neighbor ($k$NN) graph $\mathbb{G}$ on the data space. The edge weight between two connected data points is determined by the similarity between those two points. We define the adjacency matrix $\mathbf{S}$ for the graph $\mathbb{G}$ as follows: for a data point $\mathbf{X}^{(i)}$, its weight $\mathbf{S}^{(i,j)} \neq 0$ if and only if $\mathbf{X}^{(i)} \in \mathcal{N}_k(\mathbf{X}^{(j)})$ or $\mathbf{X}^{(j)} \in \mathcal{N}_k(\mathbf{X}^{(i)})$, where $\mathcal{N}_k(\mathbf{X}^{(i)})$ denotes the $k$-nearest neighborhood set for $\mathbf{X}^{(i)}$; otherwise, $\mathbf{S}^{(i,j)}=0$. In this paper, we use cosine similarity to determine nonzero weights as $\mathbf{S}^{(i,j)} = ( {\mathbf{X}^{(i)}}^T\mathbf{X}^{(j)} ) / ( \|\mathbf{X}^{(i)}\|_2\|\mathbf{X}^{(j)}\|_2 )$. The Laplacian matrix $\mathbf{L}$ of the graph $\mathbb{G}$ is defined as $\mathbf{L} = \mathbf{D} - \mathbf{S}$, where $\mathbf{D}$ is a diagonal matrix whose $i^{\mathrm{th}}$ element on the diagonal is defined as $\mathbf{D}^{(i,i)} = \sum_{j=1}^n \mathbf{S}^{(i,j)}$. With these definitions, we set up the following minimization objective for local data structure preservation:
\begin{equation*}
\label{localStruct}
\small
\mathcal{G}(\mathbf{\Theta}) = \frac{1}{2} \sum_{i=1}^n \sum_{j=1}^n \| \mathbf{Z}^{(i)} - \mathbf{Z}^{(j)} \|_2^2 \mathbf{S}^{(i,j)}  =  \rm{Tr} (\mathbf{Z}\mathbf{LZ}^T),
\end{equation*}
where $\mathbf{Z}^{(i)} = {\rm f} ( \mathbf{W_1}\mathbf{X}^{(i)} + \mathbf{b_1} )$ for $i = 1,2,\cdots,n$, and  $\mathbf{Z} = [\mathbf{Z}^{(1)}, \mathbf{Z}^{(2)}, \cdots, \mathbf{Z}^{(n)}]$. \par
The final objective function for source sample selection can be written in terms of the following minimization with respect to the parameters $\mathbf{\Theta} = [\mathbf{W_1}, \mathbf{W_2},$ $\mathbf{b_1}, \mathbf{b_2}]$ and $\mathbf{A}$:
\begin{equation}
\label{objFun}
\small
\{ \hat{\mathbf{\Theta}}, \hat{\mathbf{A}} \} = \arg\min_{\mathbf{\Theta}, \mathbf{A}} \mathcal{L}(\mathbf{\Theta}) + \mu \mathcal{C}(\mathbf{\Theta}, \mathbf{A}) + \lambda \mathcal{R}(\mathbf{A}) +  \gamma \mathcal{G}(\mathbf{\Theta}),
\end{equation}
where $\mu$, $\lambda$, and $\gamma$ are balance parameters. \par

\subsubsection{Optimization}
\label{opt}
The closed form solution of the optimization problem in  \eqref{objFun} is hard to obtain due to the $\ell_{2,1}$-norm regularization term. We employ an alternating optimization scheme to solve this problem with $\mathbf{\Theta}$ and $\mathbf{A}$ being iteratively updated, until the objective function value in  \eqref{objFun} converges or a maximum number of iterations is reached. \par
When $\mathbf{A}$ is fixed,  \eqref{objFun} becomes
\begin{equation}
\label{objFunFixA}
\hat{\mathbf{\Theta}} = \arg\min_{\mathbf{\Theta}} \mathcal{F}_1 (\mathbf{\Theta}) 
:= \arg\min_{\mathbf{\Theta}} \mathcal{L}(\mathbf{\Theta}) + \mu \mathcal{C}(\mathbf{\Theta}, \mathbf{A}) + \gamma \mathcal{G}(\mathbf{\Theta}).
\end{equation}
Following \cite{le2011optimization}, we use a limited memory Broyden-Fletcher-Goldfarb-Shanno (L-BFGS) algorithm to solve  \eqref{objFunFixA}. The L-BFGS algorithm is a type of quasi-Newton method that requires much fewer iterations to converge than first order methods such as gradient descent. Furthermore, compared with other BFGS algorithms, the L-BFGS algorithm has low computational cost, making it possible to use the whole dataset for optimization and provide more stable performance than commonly used stochastic gradient descent algorithms. For example, the dimensionality of {the} parameter $\mathbf{\Theta}$ is the sum of {the} dimensionalities of $\mathbf{W_1} \in \mathbb{R}^{m \times d}$, $\mathbf{W_2} \in \mathbb{R}^{d \times m}$, $\mathbf{b_1} \in \mathbb{R}^m$, and $\mathbf{b_2} \in \mathbb{R}^d$, which is $2md+d+m$. Compared with the conventional BFGS algorithm, which requires computing and storing of $(2md+d+m) \times (2md+d+m)$ Hessian matrices, the L-BFGS algorithm saves the past $l$ updates of $\mathbf{\Theta}$ and corresponding gradients. Therefore, denoting the number of iterations in the optimization by $t$, the corresponding time complexity of L-BFGS is $O(tlmd)$. We refer readers to~\cite{liu1989limited} for more details on L-BFGS algorithm, which we implement using the \emph{minFunc} toolbox~\cite{schmidt2005minfunc}. The solver requires the gradients of the objective function in  \eqref{objFunFixA} with respect to its parameters $\mathbf{\Theta}$. The gradients for both $\mathcal{L} (\mathbf{\Theta})$ and $\mathcal{C} (\mathbf{\Theta}, \mathbf{A})$ can be obtained through a back-propagation algorithm. We skip the details for the derivation of the gradients of both $\mathcal{L} (\mathbf{\Theta})$ and $\mathcal{C} (\mathbf{\Theta})$, which are standard in the formulation of back-propagation for an autoencoder. The resulting gradients for $\mathcal{L} (\mathbf{\Theta})$ are: 
\begin{equation*}
\small
\label{deriL}
\frac{\partial \mathcal{L} (\mathbf{\Theta})}{\partial \mathbf{W_1}} = \frac{1}{n} \mathbf{\Delta}_{\mathcal{L}2}\mathbf{X}^T, \ 
\frac{\partial \mathcal{L} (\mathbf{\Theta})}{\partial \mathbf{W_2}} = \frac{1}{n} \mathbf{\Delta}_{\mathcal{L}3}\mathbf{Y}^T, \ 
\frac{\partial \mathcal{L} (\mathbf{\Theta})}{\partial \mathbf{b_1}} = \frac{1}{n} \mathbf{\Delta}_{\mathcal{L}2}\mathbf{1}, \ 
\frac{\partial \mathcal{L} (\mathbf{\Theta})}{\partial \mathbf{b_2}} = \frac{1}{n} \mathbf{\Delta}_{\mathcal{L}3}\mathbf{1},
\end{equation*}
where each column of $\mathbf{\Delta}_{\mathcal{L}2} \in \mathbb{R}^{m \times n}$ and $\mathbf{\Delta}_{\mathcal{L}3} \in \mathbb{R}^{d \times n}$ contains the error term of the corresponding sample for the hidden layer and the output layer, respectively: 
\begin{equation*}
\small
\label{deriLdef}
\mathbf{\Delta}_{\mathcal{L}3} = ( {\rm h} (\mathbf{X}) - \mathbf{X} ) \bullet {\rm h} (\mathbf{X}) \bullet ( \mathbf{1} - {\rm h} (\mathbf{X}) ), \ \ 
\mathbf{\Delta}_{\mathcal{L}2} = ( \mathbf{W_2}^T\mathbf{\Delta}_{\mathcal{L}3} ) \bullet \mathbf{Y} \bullet ( \mathbf{1} - \mathbf{Y} ),
\end{equation*}
with $\bullet$ denoting the element-wise product operator. The gradients for $\mathcal{C} (\mathbf{\Theta}, \mathbf{A})$ are:
{\small 
\begin{equation}
\label{deriC}
\begin{aligned}
&\frac{\partial \mathcal{C} (\mathbf{\Theta}, \mathbf{A})}{\partial \mathbf{W_1}} = \frac{1}{n_{\rm trg}} \mathbf{\Delta}_{\mathcal{C}2}(\mathbf{X}_{\rm src}\mathbf{A})^T, \ 
\frac{\partial \mathcal{C} (\mathbf{\Theta}, \mathbf{A})}{\partial \mathbf{W_2}} = \frac{1}{n_{\rm trg}} \mathbf{\Delta}_{\mathcal{C}3}\mathbf{Y}_{\rm trg}^T, \\
&\frac{\partial \mathcal{C} (\mathbf{\Theta}, \mathbf{A})}{\partial \mathbf{b_1}} = \frac{1}{n_{\rm trg}} \mathbf{\Delta_{\mathcal{C}2}1}, \ \ \ \ \ \ \ \ \ \ \ \ \ \ 
\frac{\partial \mathcal{C} (\mathbf{\Theta}, \mathbf{A})}{\partial \mathbf{b_2}} = \frac{1}{n_{\rm trg}} \mathbf{\Delta_{\mathcal{C}3}1}.
\end{aligned}
\end{equation}}
Both $\mathbf{\Delta}_{\mathcal{L}2}^{(i)}$ and $\mathbf{\Delta}_{\mathcal{L}3}^{(i)}$ in  \eqref{deriL} play same roles as $\mathbf{\Delta}_{\mathcal{C}2}^{(i)}$ and $\mathbf{\Delta}_{\mathcal{C}3}^{(i)}$ in  \eqref{deriC}. Their definitions are:
\begin{equation*}
\small
\mathbf{\Delta}_{\mathcal{C}3}  = ( {\rm h} (\mathbf{X}_{\rm trg}) - \mathbf{X}_{\rm src}\mathbf{A} ) \bullet {\rm h} (\mathbf{X}_{\rm trg}) \bullet ( \mathbf{1} - {\rm h} (\mathbf{X}_{\rm trg}) ), \ \ 
\mathbf{\Delta}_{\mathcal{C}2}  = ( \mathbf{W_2}^T\mathbf{\Delta}_{\mathcal{C}3} ) \bullet \mathbf{Y}_{\rm trg} \bullet ( \mathbf{1} - \mathbf{Y}_{\rm trg} ),
\end{equation*}
where $\mathbf{Y}_{\rm trg} = {\rm f} ( \mathbf{W_1}\mathbf{X}_{\rm trg} + \mathbf{b_1} )$.
The gradients of the graph term $\mathcal{G}(\mathbf{\Theta}) = \rm{Tr} (\mathbf{YL}\mathbf{Y}^T)$ can be obtained in a straightforward fashion as follows:
\begin{equation*}
\small
\begin{aligned}
&\frac{\partial \mathcal{G} (\mathbf{\Theta})}{\partial \mathbf{W_1}}  = \frac{\partial \rm{Tr} (\mathbf{YL}\mathbf{Y}^T)}{\partial \mathbf{Y}} \cdot \frac{\partial \mathbf{Y}}{\partial \mathbf{W_1}} = 2 \left( \mathbf{YL} \bullet \mathbf{Y} \bullet ( \mathbf{1} - \mathbf{Y} ) \right) \mathbf{X}^T, \ \ \ \ \ \ 
\frac{\partial \mathcal{G} (\mathbf{\Theta})}{\partial \mathbf{W_2}}  = \mathbf{0}, \\
&\frac{\partial \mathcal{G} (\mathbf{\Theta})}{\partial \mathbf{b_1}}  = \frac{\partial \rm{Tr} (\mathbf{YL}\mathbf{Y}^T)}{\partial \mathbf{Y}} \cdot \frac{\partial \mathbf{Y}}{\partial \mathbf{b_1}} = 2 \left( \mathbf{YL} \bullet \mathbf{Y} \bullet ( \mathbf{1} - \mathbf{Y} ) \right) \mathbf{1}, \ \ \ \ \ \ \ \ \ \ 
\frac{\partial \mathcal{G} (\mathbf{\Theta})}{\partial \mathbf{b_2}} = \mathbf{0}.
\end{aligned}
\end{equation*}
To conclude, the gradients of the objective function in  \eqref{objFunFixA} with respect to $\mathbf{\Theta} = [\mathbf{W_1}, \mathbf{W_2}, \mathbf{b_1}, \mathbf{b_2}]$ can be written as
\begin{equation*}
\small
\begin{aligned}
\frac{\partial \mathcal{F}_1 (\mathbf{\Theta})}{\partial \mathbf{W_1}} & = \frac{1}{n} \mathbf{\Delta}_{\mathcal{L}2}\mathbf{X}^T + \frac{\mu}{n_{\rm trg}} \mathbf{\Delta}_{\mathcal{C}2}(\mathbf{X}_{\rm src}\mathbf{A})^T + 2\gamma \left( \mathbf{YL} \bullet \mathbf{Y} \bullet ( \mathbf{1} - \mathbf{Y} ) \right) \mathbf{X}^T, \\
\frac{\partial \mathcal{F}_1 (\mathbf{\Theta})}{\partial \mathbf{W_2}} & = \frac{1}{n} \mathbf{\Delta}_{\mathcal{L}3}\mathbf{Y}^T + \frac{\mu}{n_{\rm trg}} \mathbf{\Delta}_{\mathcal{C}3}\mathbf{Y}_{\rm trg}^T, \\
\frac{\partial \mathcal{F}_1 (\mathbf{\Theta})}{\partial \mathbf{b_1}} & = \frac{1}{n} \mathbf{\Delta_{\mathcal{L}2}1} + \frac{\mu}{n_{\rm trg}} \mathbf{\Delta_{\mathcal{C}2}1} + 2\gamma \left( \mathbf{YL} \bullet \mathbf{Y} \bullet ( \mathbf{1} - \mathbf{Y} ) \right) \mathbf{1}, \\
\frac{\partial \mathcal{F}_1 (\mathbf{\Theta})}{\partial \mathbf{b_2}} & = \frac{1}{n} \mathbf{\Delta_{\mathcal{L}3}1} + \frac{\mu}{n_{\rm trg}} \mathbf{\Delta_{\mathcal{C}3}1}.
\end{aligned}
\end{equation*}
\par
When $\mathbf{\Theta}$ is fixed,  \eqref{objFun} becomes
\begin{equation}
\label{objFunFixTheta}
\small
\begin{aligned}
\hat{\mathbf{A}} &= \arg\min_{\mathbf{A}} \mathcal{F}_2 (\mathbf{A}) 
:= \arg\min_{\mathbf{A}} \mu \mathcal{C} (\mathbf{\Theta},\mathbf{A}) + \lambda \mathcal{R} (\mathbf{A}) \\
&= \arg\min_{\mathbf{A}} \frac{\mu}{2n_{\rm trg}} \| \mathbf{X}_{\rm src}\mathbf{A} - {\rm h}(\mathbf{X}_{\rm trg};\mathbf{\Theta}) \|_F^2 + \lambda \| \mathbf{A} \|_{2,1}.
\end{aligned}
\end{equation}
Following \cite{hou2014joint}, we calculate the derivative of $\mathcal{F}_2 (\mathbf{A})$ with respect to $\mathbf{A}$ through
\begin{equation*}
\small
\frac{\partial \mathcal{F}_2 (\mathbf{A})}{\partial \mathbf{A}} = \frac{\mu}{n_{\rm trg}}[\mathbf{X}_{\rm src}^T\mathbf{X}_{\rm src}\mathbf{A} - \mathbf{X}_{\rm src}^T {\rm h} (\mathbf{X}_{\rm trg})] + \lambda \mathbf{AU},
\end{equation*}
where $\mathbf{U} \in \mathbb{R}^{n_{\rm src} \times n_{\rm src}}$ is a diagonal matrix whose $i^\textrm{th}$ element on the diagonal is 
\begin{equation}
\label{optU}
\small
\mathbf{U}^{(i,i)} = \left( 2\|\mathbf{A}_{(i)}\|_2\right)^{-1}.
\end{equation}
We add a small constant $\epsilon$ to each element in $\mathbf{A}$ to avoid overflow; thus $\|\mathbf{A}^{(i)}\|_2$ is nonzero for each $i$. In this way, $\mathcal{F}_2 (\mathbf{A})$ becomes
\begin{equation}
\label{f2U}
\small
\mathcal{S} (\mathbf{A},\mathbf{U}) = \frac{\mu}{2n_{\rm trg}} \| \mathbf{X}_{\rm src}\mathbf{A} - {\rm h}(\mathbf{X}_{\rm trg};\mathbf{\Theta}) \|_F^2 +\lambda {\rm Tr} (\mathbf{A}^T\mathbf{UA}).
\end{equation}
Therefore, when $\mathbf{U}$ is fixed, the optimal value of $\mathbf{A}$ can be obtained through 
\begin{equation}
\label{optA}
\small
\hat{\mathbf{A}} = (\mu \mathbf{X}_{\rm src}^T\mathbf{X}_{\rm src}+n_{\rm trg}\lambda \mathbf{U})^{-1} \mu \mathbf{X}_{\rm src}^T {\rm h} (\mathbf{X}_{\rm trg}).
\end{equation}
We can update $\mathbf{U}$ through  \eqref{optU} when $\mathbf{A}$ is fixed and update $\mathbf{A}$ through  \eqref{optA} when $\mathbf{U}$ is fixed with an iterative scheme until the value of $\mathcal{F}_2 (\mathbf{A})$ converges.

\subsubsection{Convergence Analysis}

The optimization of GASTL is based on an alternative scheme. When $\mathbf{A}$ is fixed, an L-BFGS algorithm is employed to optimize \eqref{objFunFixA}. The convergence analysis of L-BFGS algorithm can be found in \cite{liu1989limited}. Following \cite{hou2014joint}, we show the convergence behavior when $\mathbf{\Theta}$ is fixed as follows.
\begin{proposition}
\label{prop1}
For two variables $a$ and $b$, if $b$ is positive, then the following inequality holds:
\begin{equation}
\label{proof0}
\small
\frac{a^2}{2b}-a \geqslant \frac{b^2}{2b}-b
\end{equation}
\end{proposition}
\begin{proof}
For two arbitrary variables $a$ and $b$, we have
\begin{equation*}
\small
a^2-2ab+b^2 \geqslant 0 \Leftrightarrow a^2-2ab \geqslant b^2-2b^2
\end{equation*}
If $b$ is positive, then we get \eqref{proof0}.
\end{proof}

\begin{proposition}
\label{prop2}
When $\mathbf{\Theta}$ is fixed, the optimization procedure of \eqref{objFunFixTheta} is non-increasing over iteration by employing the optimization procedure in Section \ref{opt}.
\end{proposition}
\begin{proof}
When $\mathbf{U}$ is fixed as $\mathbf{U}^t$ in the $t^\textrm{th}$ iteration, minimizing \eqref{f2U} with respect to $\mathbf{A}$ is a convex optimization problem; thus leading to the following inequality:
\begin{equation}
\label{temp}
\begin{aligned}
&\frac{\mu}{2n_{\rm trg}} \|\mathbf{X}_{\rm src}\mathbf{A}^{t+1}-{\rm h}(\mathbf{X}_{\rm trg};\mathbf{\Theta})\|_F^2+\lambda{\rm Tr}\left[(\mathbf{A}^{t+1})^T\mathbf{U}^t\mathbf{A}^{t+1}\right] \\
\leqslant &\frac{\mu}{2n_{\rm trg}} \|\mathbf{X}_{\rm src}\mathbf{A}^t-{\rm h}(\mathbf{X}_{\rm trg};\mathbf{\Theta})\|_F^2+\lambda{\rm Tr}\left[(\mathbf{A}^t)^T\mathbf{U}^t\mathbf{A}^t\right]
\end{aligned}
\end{equation}
The trace item on the left side can be written as
\begin{equation*}
\label{temp1}
\begin{aligned}
&{\rm Tr}(((\mathbf{A}^{t+1})^T\mathbf{U}^t\mathbf{A}^{t+1})) \\
=&\|\mathbf{A}^{t+1}\|_{2,1}+{\rm Tr}(((\mathbf{A}^{t+1})^T\mathbf{U}^t\mathbf{A}^{t+1}))-\|\mathbf{A}^{t+1}\|_{2,1}\\
=&\|\mathbf{A}^{t+1}\|_{2,1}+\sum_{i=1}^{n_{\rm src}} \left( \frac{\|\mathbf{A}_{(i)}^{t+1}\|_2^2}{2\|\mathbf{A}_{(i)}^t\|_2} - \|\mathbf{A}_{(i)}^{t+1}\|_2 \right).
\end{aligned}
\end{equation*}
Likewise, the trace item on the right side can be written as
\begin{equation*}
\label{temp2}
{\rm Tr}(((\mathbf{A}^t)^T\mathbf{U}^t\mathbf{A}^t)) = \|\mathbf{A}^t\|_{2,1}+\sum_{i=1}^{n_{\rm src}} \left( \frac{\|\mathbf{A}_{(i)}^t\|_2^2}{2\|\mathbf{A}_{(i)}^t\|_2} - \|\mathbf{A}_{(i)}^t\|_2 \right).
\end{equation*}
Therefore, \eqref{temp} becomes
\begin{equation}
\label{temp4}
\begin{aligned}
&\frac{\mu}{2n_{\rm trg}} \|\mathbf{X}_{\rm src}\mathbf{A}^{t+1}-{\rm h}(\mathbf{X}_{\rm trg};\mathbf{\Theta})\|_F^2+\lambda\|\mathbf{A}^{t+1}\|_{2,1}+\lambda\sum_{i=1}^{n_{\rm src}} \left( \frac{\|\mathbf{A}_{(i)}^{t+1}\|_2^2}{2\|\mathbf{A}_{(i)}^t\|_2} - \|\mathbf{A}_{(i)}^{t+1}\|_2 \right) \\
\leqslant &\frac{\mu}{2n_{\rm trg}} \|\mathbf{X}_{\rm src}\mathbf{A}^t-{\rm h}(\mathbf{X}_{\rm trg};\mathbf{\Theta})\|_F^2+\lambda\|\mathbf{A}^t\|_{2,1}+\lambda\sum_{i=1}^{n_{\rm src}} \left( \frac{\|\mathbf{A}_{(i)}^t\|_2^2}{2\|\mathbf{A}_{(i)}^t\|_2} - \|\mathbf{A}_{(i)}^t\|_2 \right)
\end{aligned}
\end{equation}
It is easy to obtain the following inequality based on the result of Proposition \ref{prop1}:
\begin{equation}
\label{temp5}
\sum_{i=1}^{n_{\rm src}} \left( \frac{\|\mathbf{A}_{(i)}^{t+1}\|_2^2}{2\|\mathbf{A}_{(i)}^t\|_2} - \|\mathbf{A}_{(i)}^{t+1}\|_2 \right) \geqslant \sum_{i=1}^{n_{\rm src}} \left( \frac{\|\mathbf{A}_{(i)}^t\|_2^2}{2\|\mathbf{A}_{(i)}^t\|_2} - \|\mathbf{A}_{(i)}^t\|_2 \right)
\end{equation}
Plugging \eqref{temp5} into \eqref{temp4} leads to the following inequality:
\begin{equation}
\begin{aligned}
&\frac{\mu}{2n_{\rm trg}} \|\mathbf{X}_{\rm src}\mathbf{A}^{t+1}-{\rm h}(\mathbf{X}_{\rm trg};\mathbf{\Theta})\|_F^2 +\lambda\|\mathbf{A}^{t+1}\|_{2,1} \\
\leqslant &\frac{\mu}{2n_{\rm trg}} \|\mathbf{X}_{\rm src}\mathbf{A}^t-{\rm h}(\mathbf{X}_{\rm trg};\mathbf{\Theta})\|_F^2+\lambda\|\mathbf{A}^t\|_{2,1}
\end{aligned}
\end{equation}
\end{proof}

\begin{proposition}
The alternating optimization process of \eqref{objFun} with respect to $\mathbf{\Theta}$ and $\mathbf{A}$ is convergent.
\end{proposition}
\begin{proof}
When $\mathbf{A}$ is fixed as $\mathbf{A}^t$ in the $t^\textrm{th}$ iteration, an L-BFGS algorithm is used to optimize \eqref{objFun} with regard to $\mathbf{\Theta}$. Since L-BFGS algorithms are convergent,\footnote{We refer readers to \cite{liu1989limited} for more details on the convergence analysis of L-BFGS algorithms} we have 
\begin{equation}
\label{proof5}
\small
\mathcal{F} (\mathbf{\Theta}^{t+1},\mathbf{A}^t) \leqslant \mathcal{F} (\mathbf{\Theta}^t,\mathbf{A}^t).
\end{equation}
When $\mathbf{\Theta}$ is fixed as $\mathbf{\Theta}^{t+1}$ in the $(t+1)^\textrm{th}$ iteration, we can get 
\begin{equation}
\label{proof6}
\small
\mathcal{F} (\mathbf{\Theta}^{t+1},\mathbf{A}^{t+1}) \leqslant \mathcal{F} (\mathbf{\Theta}^{t+1},\mathbf{A}^t)
\end{equation}
based on Proposition \ref{prop2}. By combining \eqref{proof5} and \eqref{proof6}, we have 
\begin{equation*}
\small
\mathcal{F} (\mathbf{\Theta}^{t+1},\mathbf{A}^{t+1}) \leqslant \mathcal{F} (\mathbf{\Theta}^t,\mathbf{A}^t)
\end{equation*}
Therefore, the optimization process of \eqref{objFun} is convergent.
\end{proof}

\subsection{Classifier Training}
\label{clstrain}

The next step is to use source samples combined with target samples to train a classifier, which can then be applied to unseen samples in the target domain for classification. Since source samples have different relevance levels with the target domain, we propose a scheme to assign weights to source samples that reflect their relevance. Source domain samples with large weights are kept while others are discarded. Subsequently, a classifier is trained using both target samples and selected source samples. For each source sample, pseudo-labels that indicate the transferability of the source sample to different target classes are used as true labels during classifier training, where transferability reflects the possibility of transferring a source sample to a target domain class \cite{guo2017zero}. The transferability values of the source samples are stored in a matrix $\mathbf{Tr} \in \mathbb{R}^{n_{\rm src} \times C_{\rm trg}}$, where $C_{\rm trg}$ is the cardinality of $\mathcal{Y}_{\rm trg}$. Two pseudo-labeling schemes are proposed for comparison. Additionally, source sample weights are taken into consideration in classifier training. For each pseudo-labeling scheme, we evaluate both soft and hard classification with the softmax classifier.

\subsubsection{Source Domain Sample Reweighting}
\label{reweight}

As mentioned in Section \ref{obj}, the $\ell_2$-norm value of each row of $\mathbf{A}$ can be used to measure the relevance between the corresponding source sample and target samples. We propose a scheme to assign a weight to each source sample based on the corresponding row in $\mathbf{A}$. The weight for a source sample $\mathbf{X}_{\rm src}^{(i)} |_{i=1}^{n_{\rm src}}$ is set proportional to the $\ell_2$-norm of the corresponding row in $\mathbf{A}$. That is, for a source sample $\mathbf{X}_{\rm src}^{(i)}$, its weight vector $\mathbf{Wt} \in \mathbb{R}^{n_{\rm src}}$ has entries
\begin{equation}
\label{wt}
\mathbf{Wt} (i) = \| \mathbf{A}_{(i)} \|_2 / {\rm max}_j \left( \| \mathbf{A}_{(j)} \|_2 \right).
\end{equation}
Note that the vector $\mathbf{Wt}$ is normalized with the maximum entry value being $1$. In addition, during classifier training, all target training samples are given weight $1$.

\subsubsection{Pseudo-Labeling}
\label{pseudolabel}

We propose two transferability measure schemes and assign pseudo-labels to source samples based on transferability values.
\begin{itemize}[leftmargin=*]
\item[] \textbf{Scheme A:} For a given source sample $\mathbf{X}_{\rm src}^{(i)}$, its transferability to a target class $\mathbf{c}^{(j)}$ is measured by the square of the $\ell_2$-norm of a sub-vector consisting of the elements in the corresponding row of $\mathbf{A}$ that belong to the target samples of $\mathbf{c}^{(j)}$. That is, 
\begin{equation}
\label{tr1}
\mathbf{Tr}^{(i,\mathbf{c}^{(j)})} = \| \mathbf{A}^{(i,\mathbf{J}_{\mathbf{c}^{(j)}})} \|_2^2,
\end{equation}
where $\mathbf{J}_{\mathbf{c}^{(j)}}$ denotes the columns in $\mathbf{A}$ that correspond to class $\mathbf{c}^{(j)}$.
\item[] \textbf{Scheme B:} By following \cite{guo2017zero}, we adopt the isometric Gaussian probability \cite{socher2013zero} computed on the hidden layer representation of the trained single-layer autoencoder as the transferability of a given source sample $\mathbf{X}_{\rm src}^{(i)}$ to a target class $\mathbf{c}^{(j)}$. More concretely, the transferability is
\begin{equation}
\label{tr2}
\small
\mathbf{Tr}^{(i,\mathbf{c}^{(j)})} = \mathcal{N} ( \mathbf{Z}_{\rm src}^{(i)} | \bar{\mathbf{Z}}_{\rm trg}^{\mathbf{c}^{(j)}}, {\sigma}^2\mathbf{I} ),
\end{equation}
where $\mathbf{Z}_{\rm src}^{(i)} \in \mathbb{R}^m$ is the hidden layer representation of the source sample $\mathbf{X}_{\rm src}^{(i)}$, and $\bar{\mathbf{Z}}_{\rm trg}^{\mathbf{c}^{(j)}} \in \mathbb{R}^m$ is the mean of the hidden layer representation of target samples belonging to class $\mathbf{c}^{(j)}$. As such, the transferability is measured by the probability that the source sample belongs to a target class given the auxiliary information $\bar{\mathbf{Z}}_{\rm trg}^{\mathbf{c}^{(j)}}$.
\end{itemize}
\par
The pseudo-labels of source samples consist of a matrix $\mathbf{L} \in \mathbb{R}^{n_{\rm src} \times C_{\rm trg}}$. We assign pseudo-labels to the source samples based on their transferability values to different target domain classes. Given a source sample $\mathbf{X}_{\rm src}^{(i)}$, for a \emph{hard classifier}, we set 
\begin{equation}
\label{hardlabeling}
\mathbf{L}^{(i,j)}=
\begin{cases}
1, & j={\rm argmax}_k\mathbf{Tr}^{(i,\mathbf{c}^{(k)})}, \\
0, & {\rm otherwise}.
\end{cases}
\end{equation} 
For a \emph{soft classifier}, the normalized transferability values are used as pseudo-labels so that pseudo-labels reflect the likelihood of transferring source samples to target domain classes: 
\begin{equation}
\label{softlabeling}
\mathbf{L}^{(i,j)} = \mathbf{Tr}^{(i,\mathbf{c}^{(j)})}/\sum_{k=1}^{C_{\rm trg}} \mathbf{Tr}^{(i,\mathbf{c}^{(k)})}.
\end{equation} 
Compared with hard classifiers, soft classifiers may help improve knowledge transfer performance since they are able to capture the relationship between each single source sample and multiple target categories instead of only one category. This is especially necessary for image classification tasks since there usually exist commonalities between image categories. 

\subsubsection{Softmax Classifier Training with Weighted Samples}
\label{softmax}

We employ a softmax classifier due to its simplicity and capability to do soft classification. The training data weights are included in classifier training, which leads to the following cost function
\begin{equation}
\label{clsObjFun}
\small
{\rm J} (\mathbf{\Theta_c}) = -\frac{1}{n} \left[ \sum_{i=1}^n \mathbf{Wt}(i) \sum_{j=1}^{C_{\rm trg}} \mathbf{L}^{(i,j)} {\rm log} \frac{{\rm exp}({{\mathbf{\Theta_c}}^{(j)}}^T\mathbf{X}^{(i)})}{\sum_{l=1}^{C_{\rm trg}} {\rm exp}({{\mathbf{\Theta_c}}^{(l)}}^T\mathbf{X}^{(i)})} \right],
\end{equation}
where $\mathbf{\Theta_c} = [ \mathbf{\Theta_c}^{(1)}, \mathbf{\Theta_c}^{(2)}, \cdots, \mathbf{\Theta_c}^{(C_{\rm trg})}]$ is the classifier parameter to be optimized. We use an L-BFGS algorithm to compute the optimal value of $\mathbf{\Theta}$. The gradients needed for optimization are given by
\begin{equation*}
\small
\frac{\partial {\rm J} (\mathbf{\Theta_c})}{\partial \mathbf{\Theta_c}^{(j)}} = -\frac{1}{n} \sum_{i=1}^n \left[ \mathbf{Wt}(i) \mathbf{X}^{(i)} \left( \mathbf{L}^{(i,j)} - \frac{{\rm exp}({{\mathbf{\Theta_c}}^{(j)}}^T\mathbf{X}^{(i)})}{\sum_{l=1}^{C_{\rm trg}} {\rm exp}({{\mathbf{\Theta_c}}^{(l)}}^T\mathbf{X}^{(i)})} \right) \right].
\end{equation*}
The procedure for GASTL is summarized in Algorithm \ref{alg_wholeprocedure}.\par
\begin{algorithm}[tbp]
\caption{GASTL Algorithm}
\label{alg_wholeprocedure}
\begin{algorithmic}[1]
\Require Source dataset $\mathbf{X}_{\rm src}$; 
target dataset $\mathbf{X}_{\rm trg}$; 
graph neighborhood size $k$ ; 
autoencoder hidden layer size $m$; 
balance parameters $\lambda$ and $\gamma$.
\Ensure Softmax classifier parameter $\mathbf{{\Theta}_c}$.
\phase{Relevance Measure}
\vspace{-1.2cm}
\State Construct a $k$-nearest neighbor graph $\mathbb{G}$ on a combined dataset $\mathbf{X} = [\mathbf{X}_{\rm src} \  \mathbf{X}_{\rm trg}]$;
\State Calculate the autoencoder parameter $\mathbf{\Theta}$ and transformation matrix $\mathbf{A}$ by optimizing the objective function \eqref{objFun} with the alternating scheme described in Section \ref{opt};
\phase{Classifier Training}
\vspace{-1.2cm}
\State Calculate relevance weights for source samples according to the weighting scheme \eqref{wt};
\State Calculate the transferability of each source sample to each target class according to the transferability measure scheme \eqref{tr1} or \eqref{tr2};
\State Assign target domain class labels to source samples according to the hard pseudo-labeling scheme \eqref{hardlabeling} or the soft pseudo-labeling scheme \eqref{softlabeling};
\State Construct a softmax classifier from the target data and labels, source data, relevance weights, and pseudo-labels by optimizing the cost function \eqref{clsObjFun} to get classifier parameters $\mathbf{{\Theta}_c}$.
\end{algorithmic}  
\end{algorithm} 

\subsection{Algorithm Analysis}
In this section, we discuss the time complexity and the limitations of GASTL. \par
\emph{Time Complexity:} The time complexity for the construction of a $k$NN graph is $O(dn^2)$, where $d$ is the data dimensionality and $n = n_{\rm src} + n_{\rm trg}$. When $\mathbf{A}$ is fixed, the time complexity of using L-BFGS algorithm to optimize \eqref{objFunFixA} is $O(tlmd)$, where $t$ is the number of iterations for parameter updating and $l$ is the number of steps stored in memory; when $\mathbf{\Theta}$ is fixed, the time complexity to optimize \eqref{objFunFixTheta} is $O(n_{\rm src}^3+dn_{\rm src}^2)$, where $O(n_{\rm src}^3)$ results from matrix inversion and $O(dn_{\rm src}^2)$ results from matrix multiplication. The time complexity of source sample weighting is $O(n_{\rm src})$, which results from finding maximum value among the $\ell_2$-norm values of rows in $\mathbf{A}$. The time complexity of pseudo-labeling is $O(n_{\rm src}C_{\rm trg}^2)$. The time complexity of the softmax classifier training is $O(t_cl_cC_{\rm trg}d)$, where $t_c$ is the number of iterations for parameter updating and $l_c$ is the number of steps stored in memory in the classifier training step. Since the number of target classes $C_{\rm trg}$ is expected to be much smaller than the number of source and target samples $n_{\rm src}$ and $n_{\rm trg}$, the time complexity of GASTL is $O(dn^2+tlmd+n_{\rm src}^3+t_cl_cC_{\rm trg}d)$. \par
\emph{Limitation:} The analysis above indicates that the time complexity of GASTL is highly dependent on the number of samples, which will usually be dominated by the number of source samples. This is a potential bottleneck for GASTL to tackle with extremely large source datasets.

\section{Experiments}
\label{exp}

In this section, we evaluate the knowledge transfer performance of GASTL by comparing it with other relevant state-of-the-art transfer learning techniques. To be more specific, we first select $p$ source samples which are the most relevant to the target domain, and then use those selected source samples combined with labeled target samples to train a classifier. The classification rates and mean f1-scores on target testing samples are then used as metrics to evaluate knowledge transfer performance.

\subsection{Dataset Preparation}
\label{data}

Next, we provide information on the datasets used in our experiments. \par
\begin{itemize}[leftmargin=*]
\item {\textbf{Dataset Information: }}We employ one visual dataset (Caltech101\footnote{Dataset downloaded from \url{http://www.vision.caltech.edu/Image_Datasets/Caltech101/}. The Caltech101 dataset contains both a ``Faces`` and ``Faces easy`` class, with each consisting of different versions of the same human face images. However, the images in ``Faces`` contain more complex backgrounds. To avoid confusion between these two similar classes of images, we do not include the ``Faces easy`` images in our experiments. Therefore, we keep $100$ classes for Caltech101.}) and two natural language datasets (IMDB\footnote{Dataset downloaded from \url{https://drive.google.com/file/d/0B8yp1gOBCztyN0JaMDVoeXhHWm8/}.} and Twitter,\footnote{Dataset downloaded from \url{https://www.kaggle.com/c/twitter-sentiment-analysis2/data}} both for sentiment analysis). In order to eliminate the side effects caused by imbalanced classes, we set the number of samples from each class to be the same within each dataset through random selection. 
\item {\textbf{Feature Extraction: }}In many cases, raw data cannot be used for knowledge transfer due to possible dimensionality inconsistencies. Therefore, it is necessary to do feature extraction on each dataset to make knowledge transfer feasible. For Caltech101, we employ both the SIFTBOW feature\footnote{Codes downloaded from \url{http://files.is.tue.mpg.de/pgehler/projects/iccv09/}.} proposed in \cite{gehler2009feature} and the output of the last fully connected layer of the pre-trained VGG-19 model \cite{simonyan2015very}. For both IMDB and Twitter, we use the method from \cite{kim2014convolutional}\footnote{Codes downloaded from \url{https://github.com/yoonkim/CNN_sentence}} to do feature extraction. We denote this feature as WORD2VEC in the sequel. \par
\begin{table}[t]
\small
\caption{{Details of datasets used in our experiment.}}
\centering
\begin{tabular}{|c|p{1.1cm}<{\centering}|p{1.1cm}<{\centering}|p{1.1cm}<{\centering}|p{1.1cm}<{\centering}|} 
\hline
Dataset & Features & Samples & Classes & Type \\
\hline
Caltech101 (SIFTBOW) & 1,000 & 3,000 & 100 & Image \\
Caltech101 (VGG-19) & 4,096 & 3,000 & 100 & Image \\
IMDB & 3,000 & 6,500 & 2 & Text \\
Twitter & 3,000 & 6,500 & 2 & Text \\
\hline
\end{tabular}
\label{datasets}
\end{table} 
\item{\textbf{Source/Target Split: }}For Caltech101, we randomly separate the $100$ classes into $5$ groups with $20$ classes in each group. Five independent self-taught learning experiments were conducted on Caltech101: in each experiment samples in one group are used as target samples and those in the remaining four groups are used as source samples. For each class in the target domain, $15$ samples were used for training and $15$ samples were used for testing. For the two natural language datasets, we used one as source and the other one as target. That is, when IMDB was used as target, then Twitter was used as source and vice versa. For computational convenience, we did not use the entire datasets when either IMDB and Twitter is used as the source. We randomly selected 3,000 samples as source for both IMDB and Twitter. For each class in the target domain, 10, 100 and 1000 samples were used for training and $750$ samples were used for testing.
\end{itemize}

\subsection{Experimental Setup}

We performed classification on the target testing samples  in order to evaluate the effectiveness of the self-taught learning algorithms and two instance-based transfer methods. The three self-taught learning methods are STL \cite{raina2007self}, RDSTL \cite{wang2013robust}, and  S-Low \cite{li2018self}, introduced in Section \ref{stlintro}. The two instance-based transfer learning methods are KMM \cite{huang2007correcting} and MLS \cite{aljundi2015landmarks}. Both KMM and MLS were tailored to the scenario of self-taught learning. The computational complexity of GASTL and the five competitors are listed in Table \ref{timecomplexity}. We also compute the classification performance without knowledge transfer. \par
\begin{table}[t]
\small
\caption{Time complexity of GASTL and five competing methods.  In this table, $d$ denotes data dimensionality, $n_{\rm src}$ denotes number of source samples, $n_{\rm trg}$ denotes number of target samples, and $n=n_{\rm src}+n_{\rm trg}$, $t$ denotes number of iterations for optimization and $t_c$ denotes number of iterations for softmax classifier optimization used in GASTL and KMM and MLS that are tailored to the scenario of self-taught learning. $m$ denotes the autoencoder hidden layer size for GASTL and dictionary learning size for STL, RDSTL, and S-Low.}
\centering
\begin{tabular}{|c|c|} 
\hline
Method & Time Complexity \\
\hline
GASTL & $\mathbf{O}(dn^2+tlmd+n_{\rm src}^3++t_cl_cC_{\rm trg}d)$  \\
\hline
STL &  $\mathbf{O}(d^3+d^2n+dm)$ \\
\hline
RDSTL &  $\mathbf{O}(tnmdC_{\rm trg})$  \\
\hline
S-Low &  $\mathbf{O}(n_{\rm src}n_{\rm trg}+tnmd)$ \\
\hline
KMM & $\mathbf{O}(n^2++t_cl_cC_{\rm trg}d)$ \\
\hline
MLS & $\mathbf{O}(n++t_cl_cC_{\rm trg}d)$\\
\hline
\end{tabular}
\label{timecomplexity}
\end{table} 
Both GASTL and the compared algorithms include parameters to adjust. In this experiment, we fix some parameters and tune others through a ``grid search'' strategy. For algorithms requiring source sample selection, we select the number of source samples $p \in \{ 10, 20, 30,$ $\cdots,$ $100, 150, 200, 250,$ $\cdots,$ $500, 1000, 1500, n_{\rm src} \}$.  In GASTL, the range of hidden layer sizes is set to $m \in \{ 10, 50, 100, 200 \}$, while the balance parameters are given ranges of $\lambda \in \{ 10^{-4}, 10^{-3}, 10^{-2}, 10^{-1}, 1 \}$ and $\gamma \in \{ 0, 10^{-4}, 10^{-3}, 10^{-2}, 10^{-1} \}$. The value of $\mu$ is set to $1$. The number of nearest neighbors in a $k$NN graph is set to $5$. The value of $\sigma^2$ in \eqref{tr2} is set to $1$. For the optimization of GASTL with L-BFGS, we set the number of iterations $t_1$ and $t_2$ to be $400$ and the number of storing updates $l_1$ and $l_2$ to be $100$. \par
All three self-taught learning methods (STL, RDSTL, and S-Low) are based on dictionary learning, where the resulting sparse code vectors were used as features for classification. For these three methods, we first performed PCA on training data due to three reasons: ($i$) the features listed in Table \ref{datasets} have high dimensionalities and require large dictionary sizes, which would cause prohibitive training time; ($ii$) we noticed that the performances of these three methods were highly dependent on the choice of parameters, which means a smaller feature dimensionality would significantly reduce the training time needed due to parameter search; ($iii$) we did not see systematical differences on the performance when raw features and PCA features were employed for model learning, respectively. Following \cite{raina2007self}, we kept the number of principal components to preserve approximately $96\%$ of the training sample variance.  \par
For both KMM\footnote{Codes downloaded from \url{http://www.gatsby.ucl.ac.uk/~gretton/covariateShiftFiles/covariateShiftSoftware.html}} and MLS\footnote{Codes downloaded from \url{https://github.com/jindongwang/transferlearning/tree/master/code}}, we first obtained weights for the source samples, which are also assigned pseudo-labels. Between the two pseudo-labeling schemes, only Scheme B is applicable since Scheme A is dependent on the transformation matrix $\mathbf{A}$, and these two domain adaptation methods do not generate one. We use the features listed in Table \ref{datasets} instead of autoencoder activations as we did in GASTL. Subsequent steps were exactly the same as those for GASTL.

\subsection{Parameter Sensitivity}

We study the performance variation of GASTL with respect to the hidden layer size $m$ and the two balance parameters $\lambda$ and $\gamma$ as measured by the classification accuracy on target testing samples. We show the results on all four datasets. \par
\begin{figure*}[t]
\begin{minipage}{0.5\linewidth}
  \centerline{\includegraphics[width=4.5cm]{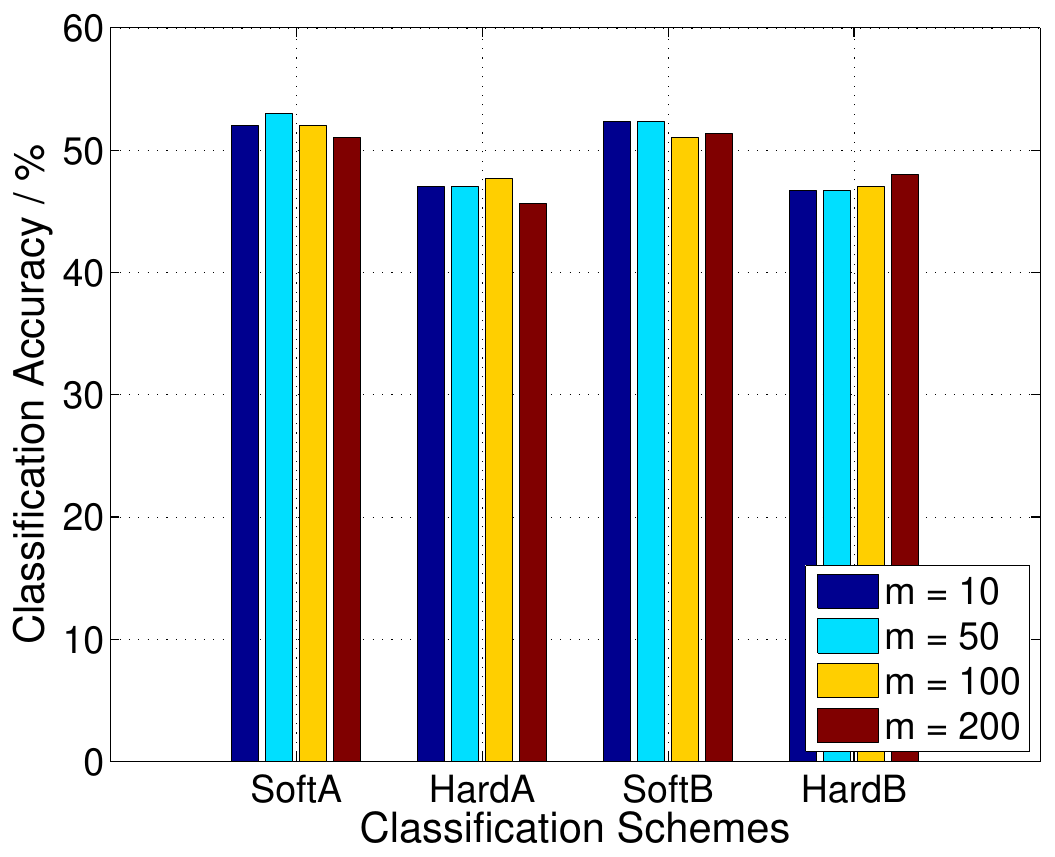}}
  \centerline{\small (a) Caltech101 (SIFTBOW)}
\end{minipage}
\hfill
\begin{minipage}{0.5\linewidth}
  \centerline{\includegraphics[width=4.5cm]{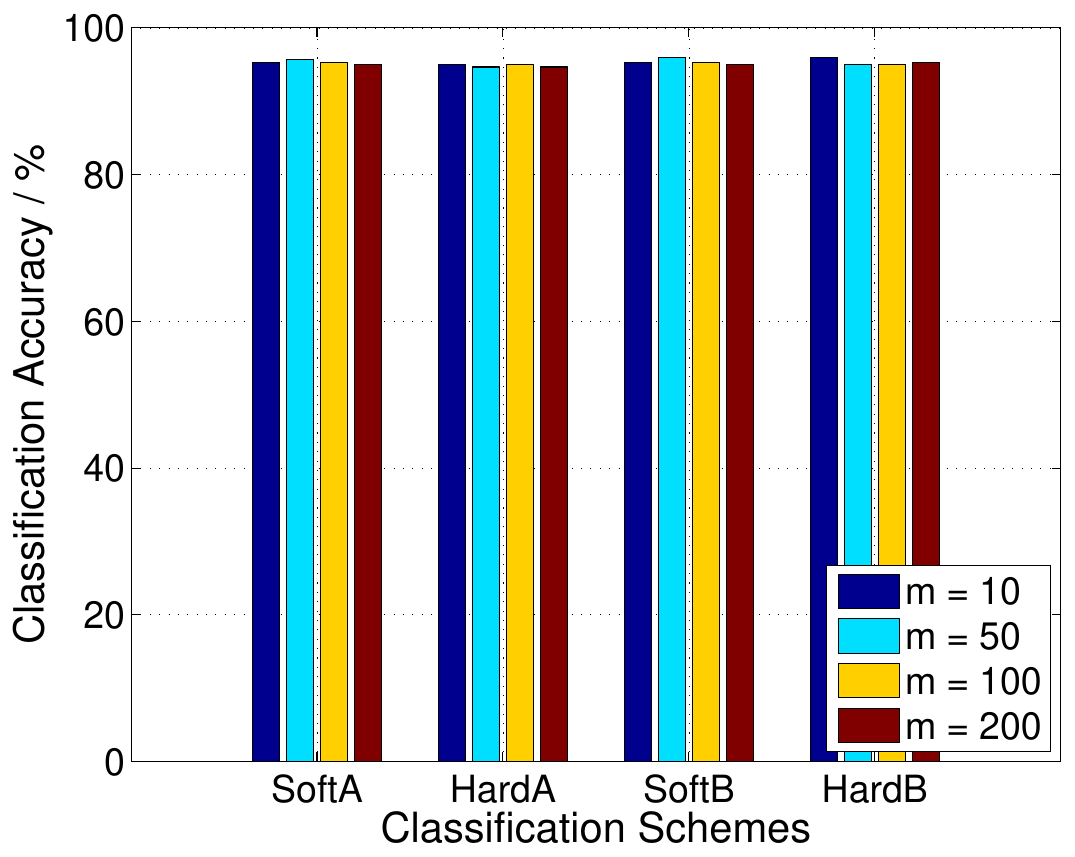}}
  \centerline{\small (b) Caltech101 (VGG19)}
\end{minipage}
\vfill
\begin{minipage}{0.5\linewidth}
  \centerline{\includegraphics[width=4.5cm]{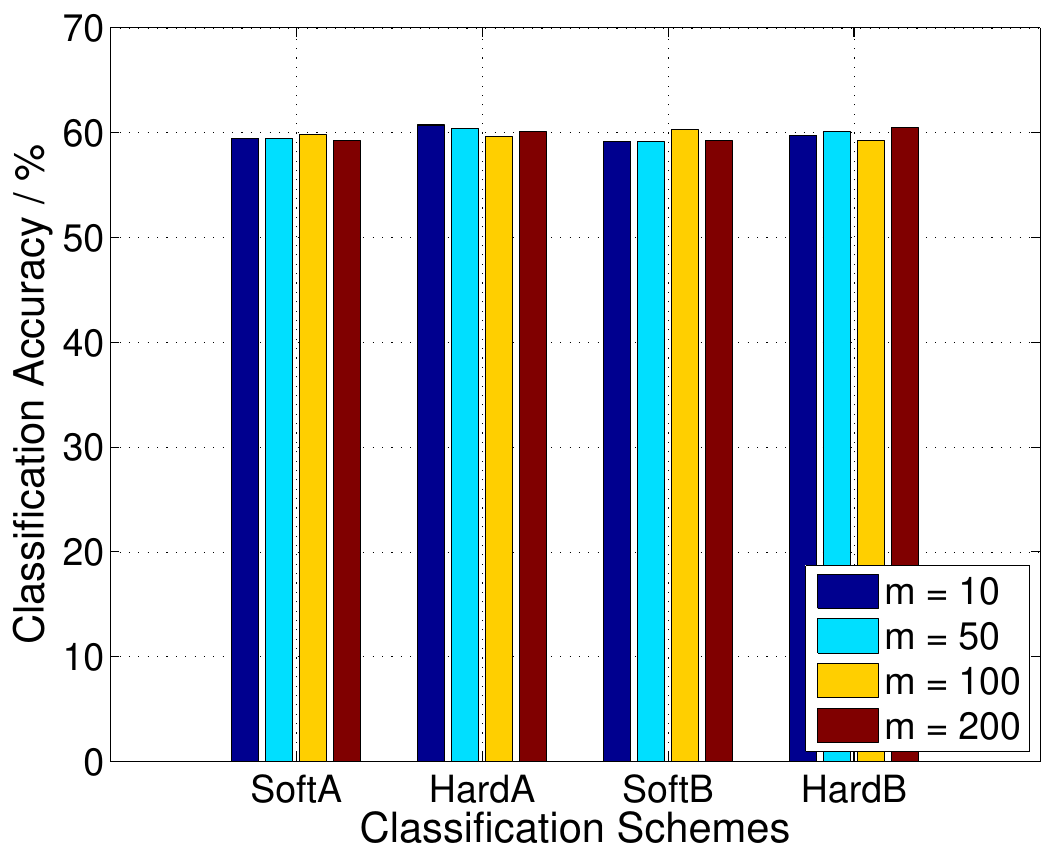}}
  \centerline{\small (c) IMDB}
\end{minipage}
\hfill
\begin{minipage}{0.5\linewidth}
  \centerline{\includegraphics[width=4.5cm]{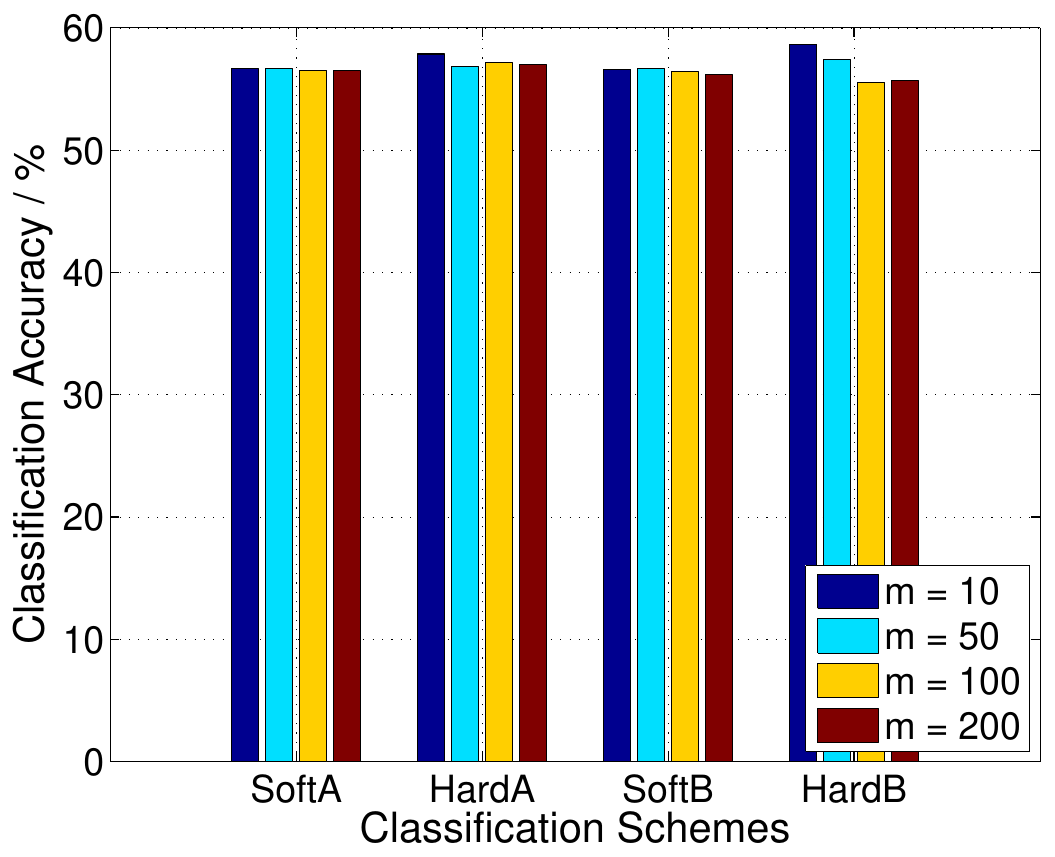}}
  \centerline{\small (d) Twitter}
\end{minipage}
\caption{Performance of GASTL in classification as a function of the hidden layer size $m$ of the autoencoder. Classification accuracy ($\%$) is used as the evaluation metric.}
\label{sensHL}
\end{figure*}

We first study the parameter sensitivity of GASTL with respect to the hidden layer size $m$. Due to limited space, we only present a small portion of our experimental results in Fig. \ref{sensHL},\footnote{For Caltech101, we use the results of one set out of five with SIFTBOW features as autoencoder inputs. For both IMDB and Twitter, we use the results with each target dataset having $10$ training samples per category.} where ``Soft'' and ``Hard'' refer to whether a soft or hard classifier is used, while ``A" and ``B" refer to the pseudo-labeling schemes. Note that when $m$ is fixed, the number of source samples used for knowledge transfer $p$ and the values of $\lambda$ and $\gamma$ are adjustable. The classification results in Fig.~\ref{sensHL} are the highest classification accuracy among all available combinations of $p$, $\lambda$, and $\gamma$ with $m$ fixed to specific values. The results show that the performance of GASTL is not too sensitive to hidden layer size on the given datasets. \par
We also study the parameter sensitivity of GASTL with respect to the balance parameters $\lambda$ and $\gamma$, under a fixed hidden layer size $m$. In order to do this, we record the classification accuracy corresponding to each combination of $\lambda$ and $\gamma$, resulting in a $5 \times 5$ matrix of performance measurement. Note that each entry in this matrix corresponds to the highest classification value over the various values of $p$ tested. We then calculate the mean and standard deviation of these $25$ elements, and the parameter sensitivity can be evaluated through the ratio between the standard deviation and the mean value. We choose the hidden layer size $m = 10$ as Fig. \ref{sensHL} shows that the performance of GASTL is not sensitive to the value of $m$. The results are listed in Table \ref{sensBalPara}, which shows that the performance of GASTL is quite stable with respect to the balance parameters $\lambda$ and $\gamma$ for Caltech101, IMDB, and Twitter. 
\begin{table}[t]
\begin{center}
\caption{Performance stability of GASTL in classification with respect to balance parameters $\lambda$ and $\gamma$. Classification accuracy mean ($\%$) and standard deviation ($\%$) are presented.}
\label{sensBalPara}
\small
\begin{tabular}{|c|c|c|c|c|}
\hline
\diagbox[width=2.5cm,height=1.5cm]{Scheme}{Dataset} & Caltech101 (S) & Caltech101 (V) & IMDB & Twitter \\
\hline
SoftA & $48.92 \pm 1.53$ & $94.48 \pm 0.47$ & $58.22 \pm 0.60$ & $55.79 \pm 0.47$ \\
\hline
HardA & $44.20 \pm 1.17$ & $92.84 \pm 1.45$ & $57.91 \pm 0.98$ & $55.41 \pm 1.13$ \\
\hline
SoftB & $48.75 \pm 1.65$ & $94.44 \pm 0.51$ & $58.08 \pm 0.54$ & $55.76 \pm 0.47$ \\
\hline
HardB & $44.27 \pm 1.45$ & $94.39 \pm 1.20$ & $57.79 \pm 0.86$ & $55.06 \pm 1.29$ \\
\hline
\end{tabular}
\end{center}
\end{table}

\subsection{Performance Comparison}

We present the classification accuracy results of GASTL and baselines on all datasets in Tables \ref{clsCaltech101siftbow} to \ref{clsnlp}, corresponding to Caltech101 (SIFTBOW), Caltech101 (VGG-19), IMDB and Twitter, respectively.\footnote{In the sequel, Caltech101 (S) and Caltech101 (V) are used to denote Caltech101 (SIFTBOW) and Caltech101 (VGG-19). Results for both IMDB and Twitter are demonstrated in Table \ref{clsnlp}} respectively. Note that ``Target Only" denotes the method that performs classifier training on target samples without knowledge transfer. In Tables \ref{clsCaltech101siftbow} and \ref{clsCaltech101vgg19}, each column corresponds to one subset, while in Table \ref{clsnlp}, each column corresponds to one training sample number in each target class. We highlight the best two performances in each experiment given that we find in many cases the best two (or even more) performance are very close to each other. \par

\begin{table}[t]
\begin{center}
\caption{Performance of GASTL and competing feature selection algorithms in classification on Caltech101 with SIFTBOW features. Classification accuracy ($\%$) is used as the evaluation metric.}
\label{clsCaltech101siftbow}
\begin{tabular}{|c|c|c|c|c|c|} 
\hline
\diagbox[width=3cm,height=1cm]{Method}{Set ID} & 1 & 2 & 3 & 4 & 5 \\
\hline
Target Only & $42.33$ & $61.67$ & $42.33$ & $48.33$ & $46.00$ \\
\hline
STL & $46.00$ & $64.33$ & $48.33$ & $46.33$ & $56.00$ \\
RDSTL & $36.67$ & $51.33$ & $37.00$ & $39.67$ & $42.00$ \\
S-Low & $35.00$ & $51.00$ & $36.67$ & $34.33$ & $36.67$ \\
\hline
KMM-Soft & $48.67$ & $66.00$ & $47.33$ & $52.67$ & $56.33$ \\
KMM-Hard & $43.67$ & $63.67$ & $43.33$ & $47.67$ & $48.33$ \\
MLS-Soft & $47.33$ & $66.33$ & $47.67$ & $51.67$ & $53.33$ \\
MLS-Hard & $45.33$ & $62.00$ & $42.00$ & $46.00$ & $47.67$ \\
\hline
GASTL-SoftA & $\bf 53.00$ & $\bf 67.67$ & $\bf 51.67$ & $\bf 56.00$ & $\bf 58.67$ \\
GASTL-HardA & $47.67$ & $64.33$ & $47.67$ & $48.67$ & $52.33$ \\
GASTL-SoftB & $\bf 52.33$ & $\bf 68.00$ & $\bf 51.33$ & $\bf 53.33$ & $\bf 58.67$ \\
GASTL-HardB & $48.00$ & $64.00$ & $46.33$ & $49.33$ & $51.00$ \\
\hline
\end{tabular}
\end{center}
\end{table}

We first provide an overall description on the comparison between GASTL and the competitors on each dataset. We can find that in Tables \ref{clsCaltech101siftbow} and \ref{clsCaltech101vgg19} the best performances are claimed by GASTL methods, and in Table \ref{clsnlp}, RDSTL is comparable to GASTL in a few cases. In other words, GASTL provides the best overall performance. We can also find that the classification performance of the two pseudo-labeling schemes are quite similar to each other in almost every case in the five tables. \par

\begin{table}[t]
\begin{center}
\caption{Performance of GASTL and competing feature selection algorithms in classification on Caltech101 with VGG19 features. Classification accuracy ($\%$) is used as the evaluation metric.}
\label{clsCaltech101vgg19}
\begin{tabular}{|c|c|c|c|c|c|} 
\hline
\diagbox[width=3cm,height=1cm]{Method}{Set ID} & 1 & 2 & 3 & 4 & 5 \\
\hline
Target Only & $94.33$ & $95.67$ & $93.00$ & $94.33$ & $93.67$ \\
\hline
STL & $95.00$ & $95.33$ & $92.00$ & $94.67$ & $94.33$ \\
RDSTL & $91.67$ & $93.00$ & $85.67$ & $90.00$ & $89.33$ \\
S-Low & $91.33$ & $90.67$ & $87.00$ & $89.67$ & $89.33$ \\
\hline
KMM-Soft & $95.33$ & $96.00$ & $93.33$ & $93.67$ & $94.67$ \\
KMM-Hard & $94.33$ & $95.33$ & $92.33$ & $94.00$ & $93.67$ \\
MLS-Soft & $94.00$ & $95.33$ & $92.67$ & $94.00$ & $93.67$ \\
MLS-Hard & $95.00$ & $95.67$ & $93.00$ & $94.33$ & $93.67$ \\
\hline
GASTL-SoftA & $95.67$ & $\bf 97.00$ & $93.67$ & $95.00$ & $\bf 96.00$ \\
GASTL-HardA & $95.00$ & $\bf 97.00$ & $\bf 94.67$ & $\bf 95.67$ & $\bf 96.00$ \\
GASTL-SoftB & $\bf 96.00$ & $\bf 97.00$ & $93.67$ & $94.67$ & $\bf 96.00$ \\
GASTL-HardB & $\bf 96.00$ & $96.67$ & $\bf 94.67$ & $\bf 95.67$ & $95.67$ \\
\hline
\end{tabular}
\end{center}
\end{table}

Our next analysis focuses on the comparison between performance generated by soft classifiers and hard classifiers. In Table \ref{clsCaltech101siftbow}, it is obvious that GASTL methods with soft classifiers provide the best performance. For image datasets such as Caltech101, it is unusual for the relevance between one source sample and a particular target class to be much larger than for other target classes. A soft classifier is able to characterize the relationship between a source sample and each target class during pseudo-labeling, while a hard classifier only selects the most similar class to each source sample and ignores other target classes, which may degrade knowledge transfer performance due to the possible useful information from other classes. This can also be validated by the results of KMM and MLS in Table \ref{clsCaltech101siftbow}, which shows the advantages of soft classifiers over hard classifiers. However, the classification rates listed in Table \ref{clsCaltech101vgg19} are quite large and similar to each other. Therefore, these results cannot provide significant information on validating the advantages of the soft classifier over the hard classifier on image datasets. On the other hand, the performance increase brought by knowledge transfer also depends on the difficulty of the classification problem. For example, the performance increase of Caltech101 using the SIFTBOW features is much larger than using the VGG-19
features. \par

\begin{table}[t]
\begin{center}
\caption{Performance of GASTL and competing feature selection algorithms in classification on IMDB and Twitter. Classification accuracy ($\%$) is used as the evaluation metric. ``TS'' denotes number of training samples in each target class. ``Twitter $\rightarrow$ IMDB'' denotes knowledge transfer from Twitter to IMDB, while ``IMDB $\rightarrow$ Twitter'' denotes the opposite direction for knowledge transfer.}
\label{clsnlp}
\begin{tabular}{|c|c|c|c|c|c|c|} 
\hline
\multirow{2}{*}{\backslashbox{Method}{TS}} & \multicolumn{3}{c|}{Twitter $\rightarrow$ IMDB} & \multicolumn{3}{c|}{IMDB $\rightarrow$ Twitter} \\ \cline{2-7} 
{} & 10 & 100 & 1000 & 10 & 100 & 1000 \\
\hline
Target Only & $57.60$ & $68.20$ & $73.27$ & $55.47$ & $58.93$ & $77.80$ \\
\hline
STL & $58.80$ & $69.53$ & $73.73$ & $53.67$ & $58.93$ & $76.93$ \\
RDSTL & $\bf 60.53$ & $\bf 70.47$ & $73.47$ & $\bf 58.60$ & $59.93$ & $63.13$ \\
S-Low & $59.93$ & $64.87$ & $73.07$ & $55.87$ & $59.47$ & $64.47$ \\
\hline
KMM-Soft & $57.13$ & $69.13$ & $73.53$  & $56.27$ & $61.20$ & $78.27$ \\
KMM-Hard & $58.00$ & $68.60$ & $73.33$ & $57.47$ & $60.33$ & $78.47$ \\
MLS-Soft & $56.40$ & $68.47$ & $72.87$  & $56.27$ & $59.40$ & $77.93$   \\
MLS-Hard & $56.80$ & $68.80$ & $72.53$ & $57.27$ & $60.40$ & $78.13$ \\
\hline
GASTL-SoftA & $59.80$ & $69.33$ & $74.60$ & $56.67$ & $61.93$ & $78.20$  \\
GASTL-HardA & $\bf 60.73$ & $\bf 70.47$ & $\bf 77.67$ & $57.87$ & $\bf 62.73$ & $\bf 78.53$ \\
GASTL-SoftB & $60.33$ & $69.27$ & $74.47$ & $56.67$ & $\bf 62.00$ & $ 78.20$ \\
GASTL-HardB & $\bf 60.53$ & $\bf 70.60$ & $\bf 77.67$ & $\bf 58.67$ & $61.93$ & $\bf 78.60$ \\
\hline
\end{tabular}
\end{center}
\end{table}

In Table \ref{clsnlp}, hard classifiers consistently provide slightly better performance than soft classifiers for GASTL methods, while for KMM and MLS, the differences are smaller. According to our experimental setup, IMDB and Twitter play interchangeable roles as source and target. In each experiment both source and target domains share a label space with two labels (``positive sentiment" and ``negative sentiment"). Therefore, in this case it is better to use hard classifier than soft classifier since the two labels indicate two mutually exclusive and largely distinguishable categories.

\subsection{Discussion}

\subsubsection{Effect of Local Data Structure Preservation}

\begin{figure*}
\begin{minipage}{0.5\linewidth}
  \centerline{\includegraphics[width=5cm]{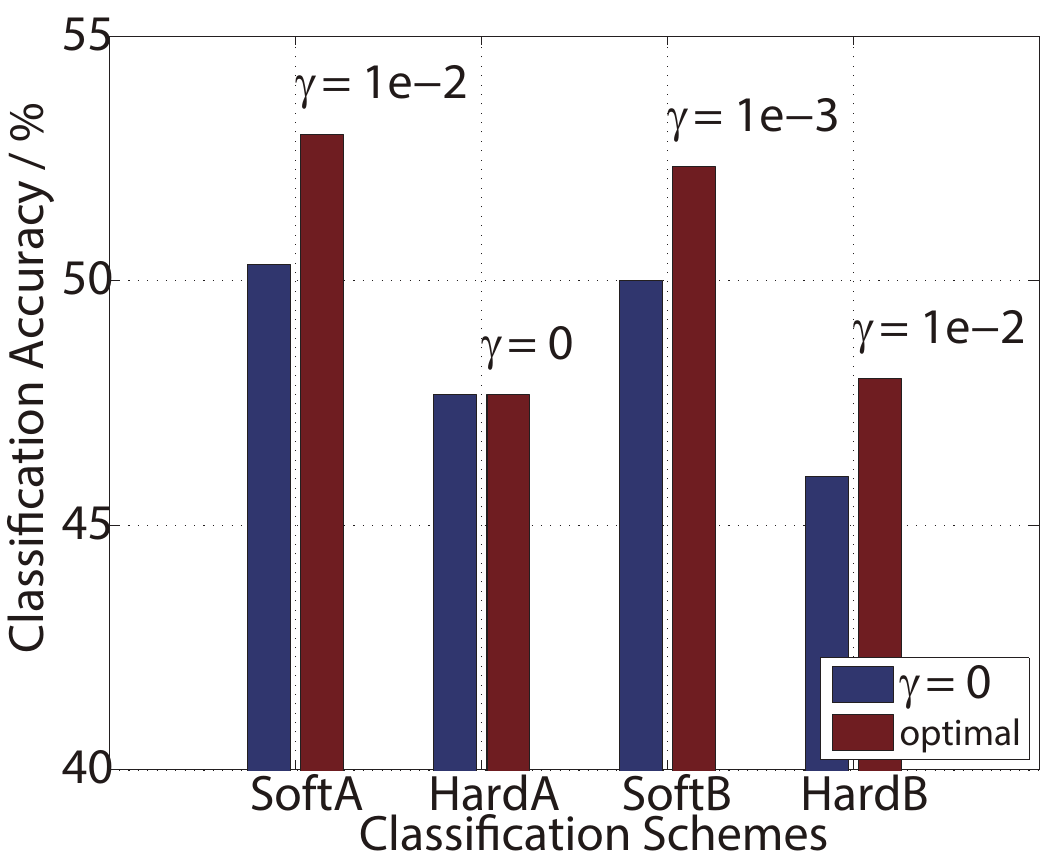}}
  \centerline{\small (a) Caltech101 (S)}
\end{minipage}
\hfill
\begin{minipage}{0.5\linewidth}
  \centerline{\includegraphics[width=5cm]{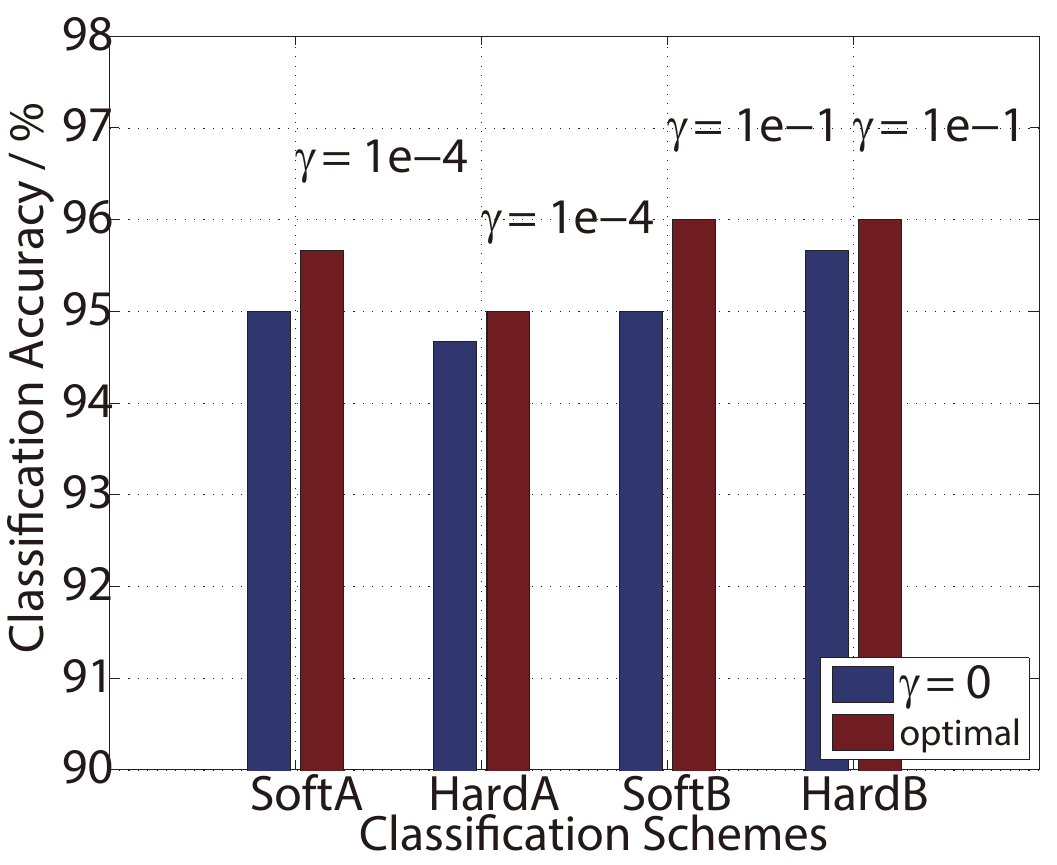}}
  \centerline{\small (b) Caltech101 (V)}
\end{minipage}
\vfill
\begin{minipage}{0.5\linewidth}
  \centerline{\includegraphics[width=5cm]{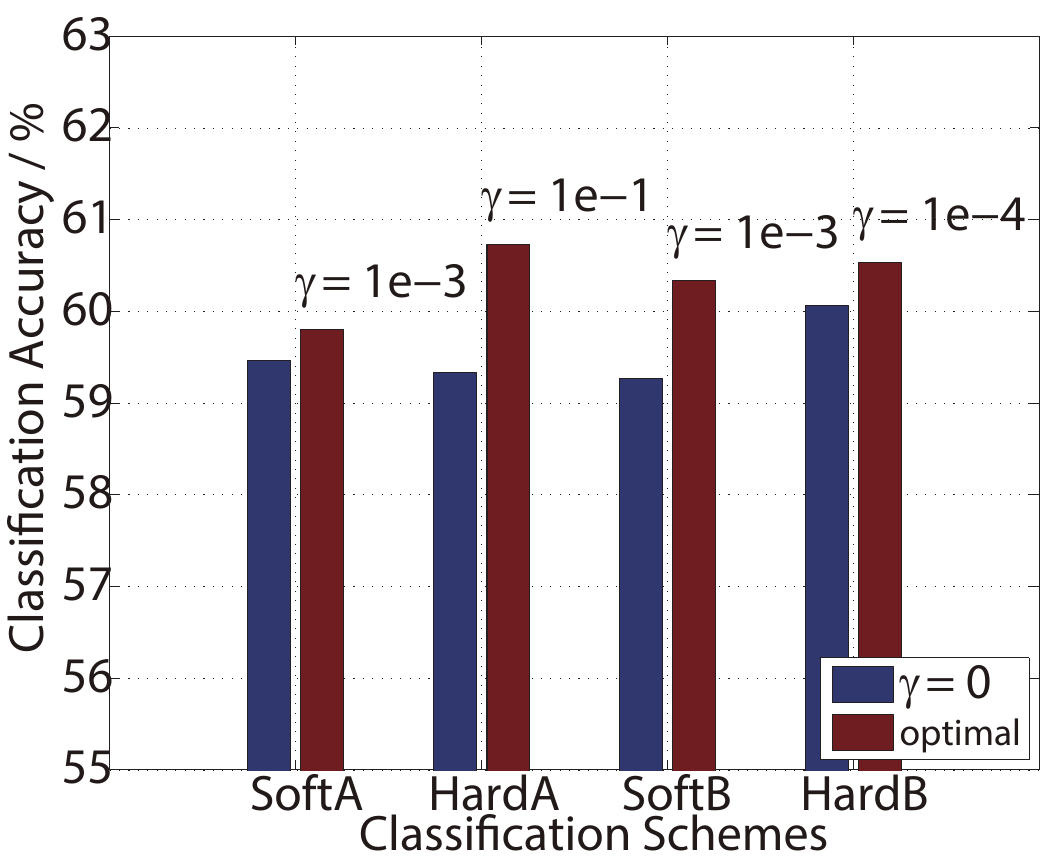}}
  \centerline{\small (c) IMDB}
\end{minipage}
\hfill
\begin{minipage}{0.5\linewidth}
  \centerline{\includegraphics[width=5cm]{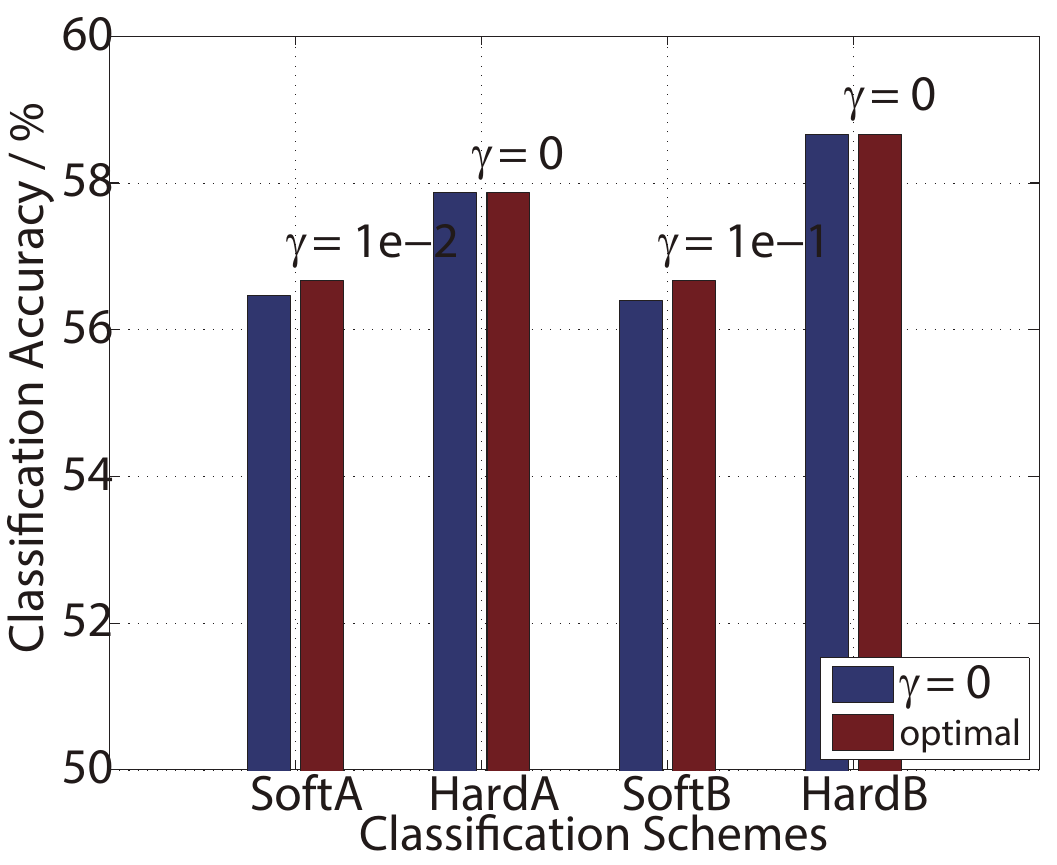}}
  \centerline{\small (d) Twitter}
\end{minipage}
\caption{Comparison of GASTL performance when $\gamma=0$ and optimal GASTL performance. Classification accuracy ($\%$) is used as the evaluation metric. Optimal values for $\gamma$ are shown.}
\label{effectGraph}
\end{figure*}

As mentioned in Section \ref{obj}, local data structure preservation provides similar representation for nearby data points. Intuitively, local data structure preservation applied in the hidden layer of the autoencoder is likely to improve knowledge transfer performance because it is able to reduce hidden layer representation distortion as it is involved in data reconstruction and pseudo-labeling. In order to measure the effect of local data structure preservation on knowledge transfer, we compare the classification performance when $\gamma=0$ with the optimal one. In Fig. \ref{effectGraph}, the comparisons on one set of Caltech101 with both SIFTBOW and VGG19 as the input to autoencoder and both IMDB and Twitter with each target dataset having $10$ training samples per category are displayed. We find that in most cases setting $\gamma=0$ cannot achieve the optimal performance. Exceptions appear in the cases of ``HardA" on Caltech101 with SIFTBOW features and both ``HardA" and ``HardB" on Twitter. Due to our observations on complete comparisons, classification rates obtained for $\gamma=0$ are predominantly lower than those obtained for $\gamma \neq 0$. Therefore, we regard these counted exceptions as outliers. Nonetheless, we found that the advantages in classification accuracy contributed by local data structure preservation are not obvious on Twitter. One possible explanation is that the local data structures in the WORD2VEC feature space cannot provide discriminative information for the Twitter dataset. Therefore, local data structure preservation negatively affected the knowledge transfer performance reflected by classification accuracy on unlabeled target samples.

\subsubsection{Effect of Source Sample Selection}

We claim that transferring knowledge from source samples indiscriminately may cause negative transfer since there is no guarantee that all source samples have sufficient relevance to the target domain. In order to demonstrate the advantage of source sample selection, we compare the classification accuracy for three different cases: the optimal value of $p$ found with GASTL, $p = 0$ (i.e.\, no transfer learning), and $p = n_{src}$ (i.e.\, no sample selection). The results shown in Fig. 3 demonstrate not only that significant performance gains are obtained via GASTL, but also that in several cases the blind consideration of all source samples can in fact result in negative transfer, as seen by the reduced performance obtained with $p = n_{src}$ versus $p = 0$.

\begin{figure*}
\begin{minipage}{0.5\linewidth}
  \centerline{\includegraphics[width=5cm]{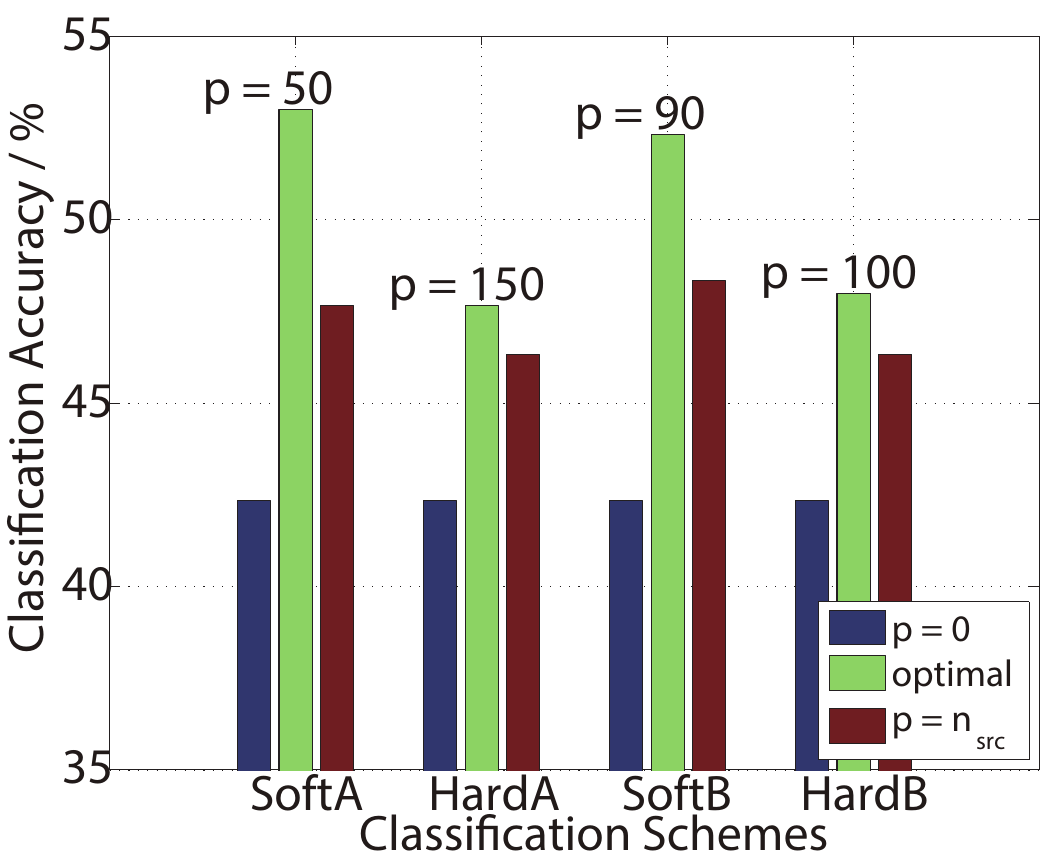}}
  \centerline{\small (a) Caltech101 (S)}
\end{minipage}
\hfill
\begin{minipage}{0.5\linewidth}
  \centerline{\includegraphics[width=5cm]{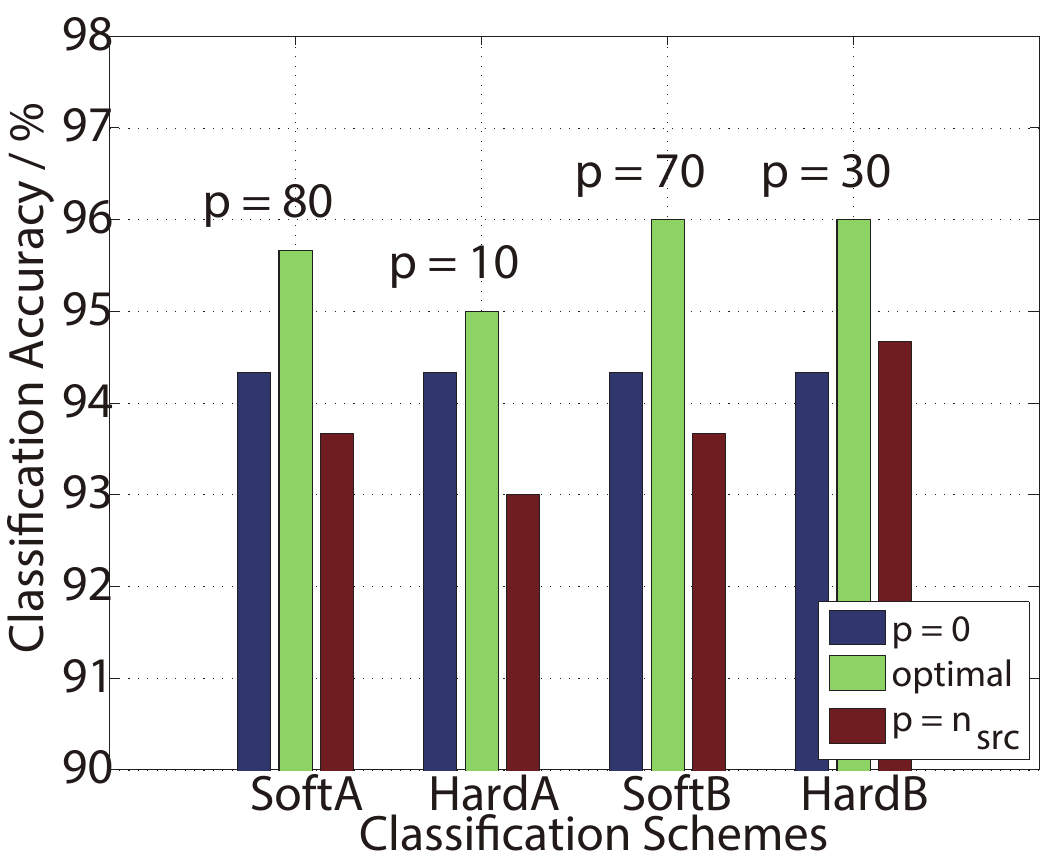}}
  \centerline{\small (b) Caltech101 (V)}
\end{minipage}
\vfill
\begin{minipage}{0.5\linewidth}
  \centerline{\includegraphics[width=5cm]{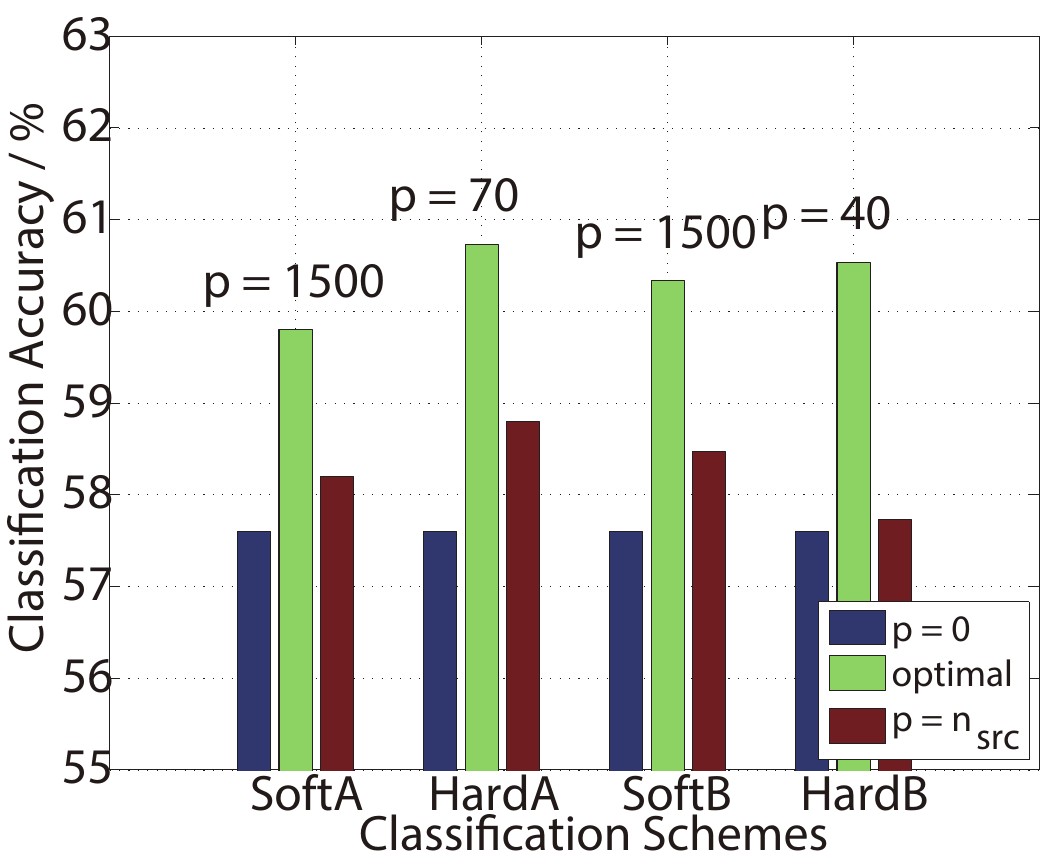}}
  \centerline{\small (c) IMDB}
\end{minipage}
\hfill
\begin{minipage}{0.5\linewidth}
  \centerline{\includegraphics[width=5cm]{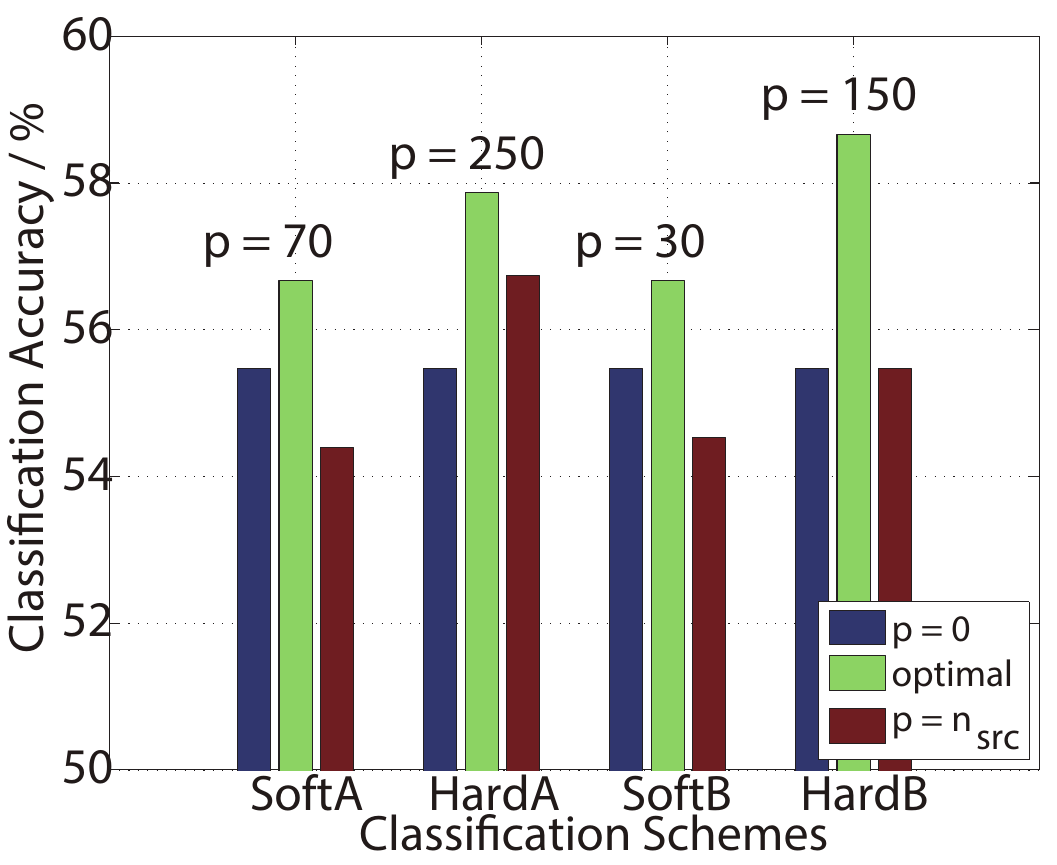}}
  \centerline{\small (d) Twitter}
\end{minipage}
\caption{Comparison of GASTL performance when all source samples are used for classifier training and optimal GASTL performance. Classification accuracy ($\%$) is used as the evaluation metric. Optimal values for $p$ are shown.}
\label{effectSampSelect}
\end{figure*}

\section{Conclusions and Future Work}
\label{conclusion}

In this paper, we propose a graph and autoencoder based self-taught learning (GASTL) method. The main innovations in our self-taught learning methodology with respect to the literature can be summarized as (a) leveraging relevance metrics to select a subset of source samples in transfer learning; (b) considering cross-domain relevance for classifier training; and (c) developing our method for hard as well as soft classification problems.  With our proposed framework, we decrease negative transfer and improve knowledge transfer performance in many scenarios. Experimental results demonstrate the advantages of GASTL versus methods in the literature. \par
For future work, we will focus on extending our work from a single-layer autoencoder-based design to one based on deep neural networks \cite{yu2016deep, hong2015multimodal} as well as reducing time complexity. We also plan to integrate discriminative information of target samples into our framework.

\section*{Acknowledgment}

We thank Dr. Sheng Li for providing the code of paper \cite{li2018self}.\par
This research is supported in part by the Nanyang Assistant Professorship (NAP); AISG-GC-2019-003; NRF-NRFI05-2019-0002; NTU-SDU-CFAIR (NSC-2019-011); and NTU-WeBank JRI (NWJ-2019-007).

\end{document}